\documentclass[lettersize,journal]{IEEEtran}

\usepackage{epsfig} 
\usepackage{amsmath} 
\usepackage{amssymb}  

\usepackage{bm}
\usepackage{caption}
\usepackage{subcaption}
\usepackage[hidelinks]{hyperref}

\renewcommand{\baselinestretch}{0.97}

\newcommand{\va}{\mathbf a}

\newcommand{\vf}{\mathbf f}

\newcommand{\vv}{\mathbf v}

\title{\LARGE \bf
	Quadratic Programming-based Reference Spreading Control for \\
	Dual-Arm Robotic Manipulation with Planned Simultaneous Impacts
}

\author{Jari van Steen, Gijs van den Brandt, Nathan van de Wouw, Jens Kober and Alessandro Saccon
 \thanks{This work was partially supported by the Research Project I.AM. through the European Union H2020 program under GA 871899 and the EFRD OP Zuid project ‘Holland Robotics Logistiek’, ref. PROJ-02535.}
 \thanks{Jari van Steen, Gijs van den Brandt, Nathan van de Wouw and Alessandro Saccon are with the Faculty of Mechanical Engineering, Eindhoven
University of Technology, 5612 AE Eindhoven, The Netherlands (e-mail: j.j.v.steen@tue.nl, a.a.h.m.v.d.brandt@tue.nl, n.v.d.wouw@tue.nl, a.saccon@tue.nl)}
 \thanks{Jens Kober is with the Department of Cognitive Robotics, Delft University of Technology, 2628 CN Delft, The Netherlands (e-mail: j.kober@tudelft.nl)}
}

\begin{document}

	\maketitle
	\thispagestyle{empty}
	\pagestyle{empty}

	\begin{abstract}
		
		With the aim of further enabling the exploitation of intentional impacts in robotic manipulation, a control framework is presented that directly tackles the challenges posed by tracking control of robotic manipulators that are tasked to perform nominally simultaneous impacts. This framework is an extension of the reference spreading control framework, in which overlapping ante- and post-impact references that are consistent with impact dynamics are defined. In this work, such a reference is constructed starting from a teleoperation-based approach. By using the corresponding ante- and post-impact control modes in the scope of a quadratic programming control approach, peaking of the velocity error and control inputs due to impacts is avoided while maintaining high tracking performance. With the inclusion of a novel interim mode, we aim to also avoid input peaks and steps when uncertainty in the environment causes a series of unplanned single impacts to occur rather than the planned simultaneous impact. This work in particular presents for the first time an experimental evaluation of reference spreading control on a robotic setup, showcasing its robustness against uncertainty in the environment compared to three baseline control approaches.

		
	\end{abstract}
	 
	\section{Introduction}\label{sec:introduction} 

	We as humans are naturally skilled in exploiting impacts to perform or speed up the execution of contact tasks. Examples of this include activities like running or kicking a ball, or more everyday tasks like opening a door or grabbing objects. While straightforward for us, exploitation of impacts is challenging to translate to the world of robotics. State-of-the-art robot control approaches such as \cite{Salehian2018} enforce almost-zero relative velocity between the end effector and the object or environment when establishing contact, avoiding impacts altogether. While establishing contact at zero speed is viable and sometimes necessary, there are various applications where exploiting impacts in a human-like way can be beneficial. 
	The exploitation of impacts has been and is still an active area of research in robot locomotion \cite{Badri-Sprowitz2022,Katz2019,Westervelt2002}, while it is now starting to be explored also in robot manipulation with objects of non-negligible weight \cite{Stouraitis2020,Khurana2021}.

	One particular example of an industrial use case that would benefit from exploitation of impacts concerns swift pick-and-place operations using robotic dual-arm manipulation \cite{Dehio2022, Bombile2022} in logistic scenarios such as palletization or depalletization. Utilization of impacts in dual-arm manipulation in a human-like way can decrease cycle times in such operations. However, several challenges arise when dealing with robotic impact-aware manipulation. 
 
    \begin{figure}
		\centering
		\includegraphics[width=\linewidth]{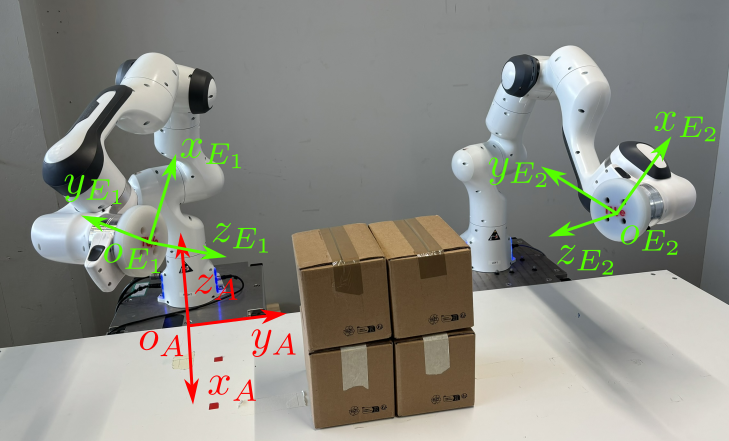}
		\caption{Depiction of the dual-arm robotic setup used for experimental validation of the presented control approach.}
		\label{fig:setup}
	\end{figure}	

    \subsection{Challenges in impact-aware manipulation}\label{sec:challenges}

    While the field of impact-aware manipulation holds too many challenges to all address in this work
    , some key challenges that are relevant to this endeavour are presented here.  

    \subsubsection{Hardware limit violation}
	First of all, hardware damage as a result of excessive external forces needs to be avoided. This can be done by using suitable hardware solutions, for example series elastic actuators \cite{Pratt1995}, variable impedance actuators \cite{Vanderborght2013}, series viscoelastic actuators \cite{Parietti2011}, compliant end effectors \cite{Dehio2022}, or an overload clutch in the drivetrain \cite{Ostyn2021}. Other approaches focus on control-based solutions, limiting ante-impact velocities to avoid such violation of hardware limits \cite{Wang2019}. 
    
    \subsubsection{Impact-consistent reference generation} Another challenge is related to motion planning, as tracking control approaches require references that are consistent with the impact dynamics to avoid steps in the velocity tracking error. These steps can result from impact-induced jumps in the joint velocities, and can translate to undesired steps in the actuator input torques, potentially damaging or destabilizing the system, and increasing energy consumption. One way to formulate such impact-consistent references is by using an analytical impact map \cite{Glocker2006} to generate matching ante- and post-impact references \cite{Steen2022}. However, this impact map formulation does not translate easily to complex impact scenarios of one or multiple nDOF robots, such as the scenario considered in this work. Alternatively, one could consider approaches based on realistic simulations, which do require appropriate software. Another possibility, which will be considered in this work, is the use of demonstrations on real hardware \cite{Mulling2013} to generate motions containing impacts and extract suitable references. 
        
    \subsubsection{Error peaking} The main challenge addressed in this work is the avoidance of so-called error peaking. Under traditional tracking control, impact-induced jumps in the joint velocities will cause the velocity tracking error to peak around the impact time as a result of an inevitable mismatch between the actual impact time and the nominal impact time \cite{Biemond2013,Forni2013,Leine2008}. This peak in the tracking error can result in similar peaks in the input torques, potentially inducing additional vibrations, or damage/destabilize the system and increase energy consumption as described for impact-induced input steps. 
	An additional challenge arises when several impacts are planned to be executed simultaneously \cite{Rijnen2019} as is the case for humanoid jumping, manipulation tasks with impacts between significantly large surfaces, or dual-arm manipulation. Inevitable differences in the impact timing between different bodies, or misalignment of the impacting surface(s) of the robot and the object leads to multiple separate impacts to take place. This results in uncertainty in the contact states during this impact sequence, complicating tracking control when compared with a single impact scenario. The goal of this paper is hence to enable tracking control in mechanical systems that experience nominally simultaneous impacts at multiple contact points.

    \subsection{Related work on impact-aware tracking control}

    Several works have been devoted to tackle the issues related to error peaking that arise when dealing with single or simultaneous impacts in the context of tracking control. This includes \cite{Biemond2013}, which resolves the problems with error peaking by re-defining the tracking error using a distance function that takes the predicted velocity jumps into account, such that possible discrepancies between the predicted and actual impact times do not affect the tracking error.  
    Alternatively, \cite{Yang2021} removes peaks in the velocity error by projecting the control objectives onto a subspace that is invariant to the impact event, essentially removing velocity feedback around the impact event for all control objectives affected by the impact, while retaining full control over all other control objectives.     
    In \cite{Morarescu2010}, a so-called transition phase is introduced to apply tracking control during contact transitions, which is proven to result in a stable closed-loop system. 
    A heuristic approach to deal with contact transitions is proposed in \cite{Mason2016}, which blends the gains from an ante-impact LQR controller to those of a post-impact LQR controller as soon as an impact is detected, in order to remove discontinuities in the joint torques.

    Another framework that was designed to tackle the issues related to error peaking, which will be extended in this work, is the framework of reference spreading (RS). RS was introduced in \cite{Saccon2014} and has been further developed in \cite{Rijnen2015,Rijnen2017} for single impact events, with \cite{Rijnen2019a} providing a formal proof of local exponential stability for motions with isolated single impacts. This framework enables tracking control of mechanical systems through a hybrid control approach with an ante- and post-impact reference that overlaps about the nominal impact time. Switching from the ante- to the post-impact control mode is done based on detection of the impact, instead of being tied to the nominal impact time, to remove the peak in the velocity error.

    While RS control was initially defined for single impact events, an approach to deal with nominally simultaneous impacts has been introduced in \cite{Rijnen2019}. Uncertainty in the environment could cause a sequence of impacts rather than the planned nominally simultaneous impact, resulting in unexpected velocity jumps and a contact state that neither corresponds to the ante-impact nor the post-impact state. To avoid peaks and steps in the actuator inputs that would occur when naively applying velocity feedback, an additional interim-impact mode was defined.    
    This interim mode is active in the time frame when contact between two impacting bodies has only been partially established, and uses a feedforward signal based on the ante-impact mode without any position or velocity feedback to be independent from any jumps in velocity. 
	In \cite{Steen2022}, an extension to the RS framework was proposed by defining a novel interim mode, which also uses position feedback during this interim mode to pursue persistent contact establishment without relying on velocity feedback. Furthermore, the approach in \cite{Steen2022} is cast into the quadratic programming (QP) robot control framework \cite{Salini2011,Cisneros2018,Bouyarmane2019}, where the control input is obtained from a linearly constrained quadratic optimization problem. QP control can be used for control of robots with multiple tasks in joint space or operational space, under a set of constraints ensuring, for example, adherence to joint limits or avoidance of unwanted collisions. This is essential in real-life robotic applications, and is required to eventually extend the RS framework to other more complex robotic systems, such as humanoids or quadrupeds.  
    A control algorithm based on RS using time-invariant references rather than the traditionally used time-based references has been developed in \cite{Steen2022a,Steen2022b}, also using a QP control framework. While experimental validation for RS has been provided for a 1 degree of freedom (DOF) mechanical system with single impact events in \cite{Rijnen2020}, RS for systems with simultaneous impacts, or for nDOF robots with $n>1$ in general, has not yet been realised.

    \subsection{Contribution}

    In this work, a QP-based control approach is presented that is used to prevent error peaking in robot control for motions that include nominally simultaneous impacts. This approach uses the aforementioned RS framework with an ante-impact, interim and post-impact control mode to prevent unwanted peaks and jumps in the velocity tracking error and thus in the actuator input signals, even when uncertainty causes a mismatch between the actual and predicted impact timing. The interim mode in particular is designed to achieve full sustained contact without input peaks and steps when this uncertainty additionally causes an unexpected loss of impact simultaneity. The approach is validated using a real-life setup consisting of two 7DOF Franka Emika robots in a dual-arm grasping scenario as shown in Figure \ref{fig:setup}.
    
    This paper is an extension of \cite{Steen2022}, which validated a similar RS control approach for simultaneous impacts using numerical simulations of a 3DOF planar manipulator. This work extends that numerical validation to an experimental validation, which marks the first time that the RS control framework is validated using a physical robot setup containing more than a single DOF.     
    Furthermore, compared to the one in \cite{Steen2022}, a novel interim mode is defined. The interim mode is in both approaches entered upon detection of the first impact, but given the practical challenges in determining when sustained full contact is established on a real setup, it was now decided to switch to the post-impact mode after a pre-defined time duration, assuming the impact is completed in this time frame. The novel interim mode design ensures a smooth transition not only from the ante-impact to the interim mode, but also from the interim mode to the post-impact mode, avoiding unwanted input steps even when the system resides in the interim mode for a longer time than strictly necessary.

    Additional effort was put into resolving the first two challenges mentioned in Section \ref{sec:challenges} that arise with validation of RS on a physical dual-arm setup (violation of actuator limits and impact-consistent reference generation). 
    While the proposed control approach limits peaks in the control inputs in response to impact-induced velocity jumps, the physical impact event will still cause a peak in the external force, potentially leading to hardware failure or joint limit violations. 
    To mitigate these problems, a custom compliant silicone end effector was created and used, reducing such external force peaks. 
    A reference consistent with the impact dynamics is generated by demonstrating the task at hand on the real robotic setup using a teleoperation framework with impedance control, see Figure \ref{fig:teleoperation}. This demonstration provides insight into the impact dynamics for the task at hand without requiring explicit knowledge of impact models, and can hence be used as starting point for the reference used in autonomous execution of the desired motion. 
    Since a human operator can modify his reference accordingly to compensate for poor tracking, low control gains can be used during teleoperation. This heavily reduces the effect of input peaks and steps resulting from the impact-induced velocity jumps. 
    As autonomous control cannot make use of a user that closes the loop with vision, high gains are required to accurately execute a desired motion, which motivates the use of the proposed RS framework. Please note that the teleoperation-based approach for generation of an impact-consistent reference can be replaced by an autonomous impact-aware motion planner without requiring any modifications to the presented control approach. 
    

    \subsection{Outline}
 
	This paper is structured as follows. Section \ref{sec:robot_dynamics} presents the adopted notation together with the equations of motion of the setup that will be used to demonstrate the control approach. In Section \ref{sec:reference_generation}, the teleoperation-based approach used to generate a reference consistent with the impact dynamics is described, as well as the reference extension procedure to formulate extended ante- and post-impact references used in RS. In Section \ref{sec:control_approach}, the control approach itself is formulated. An experimental validation of the proposed approach is finally presented in Section \ref{sec:validation}, followed by the conclusions in Section \ref{sec:conclusion}. A video highlighting the experimental validation is additionally supplied.





	
		
	
	
	\section{Notation and robot dynamics}\label{sec:robot_dynamics}

    While the approach proposed in this work is broadly applicable for robot control in different scenarios with (ideally simultaneous) impacts, the dual 7DOF robot-arm setup depicted in Figure \ref{fig:setup} is used throughout this paper to illustrate and demonstrate the approach. Both robots contain a silicone end effector, with a frame $E_i$, $i \in \{1,2\}$, attached. Using a shortened variant of the notation of \cite{Traversaro2019}, we denote the position and rotation of the end effector frames $E_i$ with respect to the inertial frame $A$ as $\bm p_i := {}^A \bm p_{E_i}$ and $\bm R_i := {}^A \bm R_{E_i}$, respectively. We denote its twist as $\vv_i = \left[\bm v_{i}, \bm \omega_i\right]  := {}^{E_i[A]} \vv_{A,E_i}$, and its acceleration as $\va_i = \left[\bm a_{i}, \bm \alpha_i\right]  := {}^{E_i[A]} \va_{A,E_i}$ with linear and angular velocity, and linear and angular accelerations $\bm v_{i}, \bm \omega_{i}, \bm a_{i}, \bm \alpha_{i}$, respectively. 
	
	Each robot contains 7 actuated joints, which are assumed to be rigid, with joint displacements $\bm q_i$ for robot index $i\in \{1,2\}$. The end effector twists and accelerations can be expressed in terms of the joint velocities and accelerations as
	\begin{equation}
	\vv_i = \bm J_i(\bm q) \dot{\bm{q}}_i,
	\end{equation}
        \begin{equation}
	\va_i = \bm J_i(\bm q) \ddot{\bm{q}}_i + \dot{\bm J}_i(\bm q, \dot{\bm q}) \dot{\bm{q}}_i 
	\end{equation}
	with geometric Jacobian $\bm J_i(\bm q)$.     
	The equations of motion of both robots are given by 
	\begin{equation}\label{eq:eom}
	\bm M_i(\bm q_i)\ddot{\bm q}_i + \bm h_i(\bm q_i,\dot{\bm q}_i)= \bm \tau_i + \bm J_{i}^T(\bm q_i)  \mathbf{f}_i 
	\end{equation} 
	with  mass matrix $\bm M_i(\bm q_i)$, vector of gravity, centrifugal and Coriolis terms $\bm h_i(\bm q_i,\dot{\bm q}_i)$, applied joint torques $\bm \tau_i$, and contact wrench $\vf_i$. For ease of notation, the explicit dependency on $\bm q_i$ (or $\dot{\bm q}_i$) is dropped for the remainder of the document.

	\section{Extended reference generation via teleoperation} \label{sec:reference_generation}
	
	At the core of RS lies the formulation of ante-impact and post-impact references that are consistent with the impact dynamics and overlap around the impact time. 
	In this section, first, a methodology for obtaining ante- and post-impact references that match the impact dynamics is described, which uses teleoperation with a QP-based impedance controller. We reiterate that the proposed RS control approach is not tied to this teleoperation approach, and can be replaced by an autonomous impact-aware motion planner. Subsequently, a methodology for extending the obtained reference around the nominal impact time is described. This allows for a formulation of the tracking error that does not contain impact-induced peaks or jumps which would translate to the input torques, as mentioned in Section \ref{sec:introduction}.

	\subsection{Teleoperation controller}\label{sec:tele_impedance}
	
 
	\begin{figure}
		\centering
		\includegraphics[width=\linewidth]{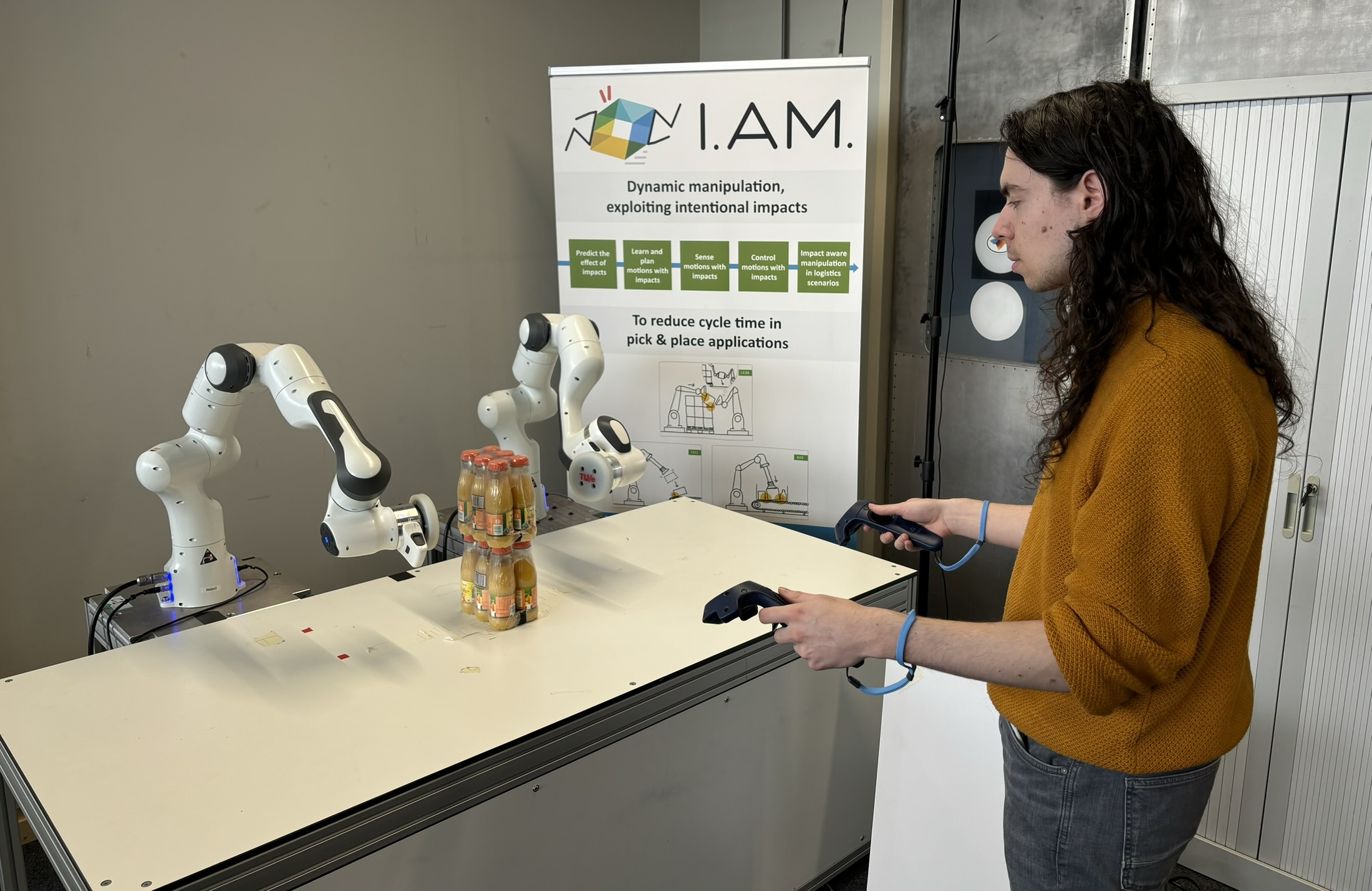}
		\caption{Depiction of the teleoperation procedure, where a user uses the VIVE handheld controller devices to prescribe a reference for both end effectors.}
		\label{fig:teleoperation}
	\end{figure}
    \begin{figure*}
		\centering
		\includegraphics[trim={2cm 23.0cm 5.3cm 0.7cm}, clip]{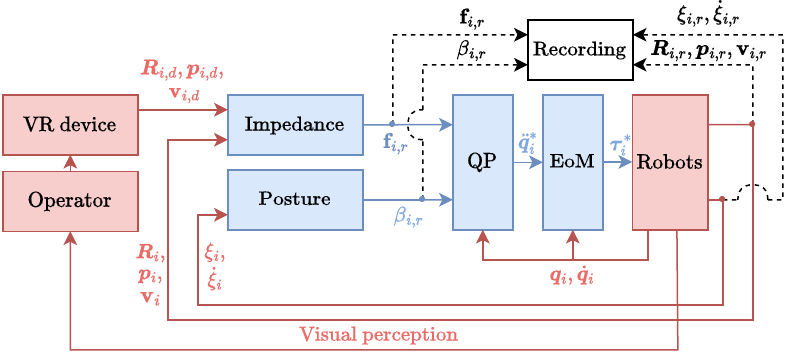}
		\caption{Visualization of the robot control scheme based on QP robot control employed for recording a reference through teleoperation. All physical entities are red, the controller-related signals are blue, and the recorded signals are black.}
		\label{fig:flow_rec}
	\end{figure*}

    The teleoperation interface in this work consists of two HTC VIVE handheld controller devices\footnote{VIVE handheld controllers: \href{https://www.vive.com/eu/accessory/controller/}{https://www.vive.com/eu/accessory/controller/}.}, which are used in combination with the openVR SDK\footnote{openVR SDK: \href{https://github.com/ValveSoftware/openvr}{https://github.com/ValveSoftware/openvr}.}. By tracking the movements of the handheld devices, a reference for the end effector position, orientation and twist of both robots is obtained, as depicted in Figure \ref{fig:teleoperation}. These desired references are described over time $t$ with subscript $d$ for a robot with index $i\in \{1,2\}$ as $\bm p_{i,d}(t)$, $\bm R_{i,d}(t)$, and $\vv_{i,d}(t)$, corresponding to $\bm p_{i}$, $\bm R_{i}$, $\vv_{i}$, respectively, as described in Section \ref{sec:robot_dynamics}.

	During teleoperation, a QP control approach \cite{Bouyarmane2019,Salini2010} is used to generate a control action that drives the end effectors to the prescribed reference. Using QP control carries the benefit of allowing to pursue multiple tasks while enforcing constraints that can prevent e.g. joint limit violations, as further emphasized in Section \ref{sec:introduction}. 
    A visualization of the different steps in the control approach used for teleoperation, including which signals are recorded, is shown in Figure \ref{fig:flow_rec}, with further details provided in the remainder of this section. Specifically, a QP with optimization variables $\ddot{\bm q}_i$ with $i\in \{1,2\}$ is solved online at fixed time intervals $\Delta t$ to determine a suitable desired joint acceleration $\ddot{\bm q}^*_i$ for both robots. This acceleration is converted into a desired joint torque $\bm \tau^*_i$ by using the equations of motion of the robot in free motion, i.e. 
	\begin{equation}\label{eq:EOM_free}
	\bm \tau^*_i = \bm M_i\ddot{\bm q}^*_i + \bm h_i.
	\end{equation}
    To limit the number of optimization variables and to comply with the QP solver of the mc\_rtc software framework used for experimental validation, $\ddot{\bm q}^*_i$ is converted to $\bm \tau^*_i$ using \eqref{eq:EOM_free} after solving the QP rather than including both $\ddot{\bm q}^*_i$ and $\bm \tau_i$ as optimization variables in the QP itself.  This $\bm \tau^*_i$ is sent as joint torque reference to the low-level torque control scheme built into the Franka Emika robots, with additional friction compensation provided according to the approach described in \cite{Gaz2019}. 
	
	The cost function of the QP is described by means of a weighted sum of the costs corresponding to different tasks that are to be executed. By minimizing this cost function under a set of yet-to-be-defined constraints, all tasks are executed to the maximum extent possible without violation of these constraints. The tasks and constraints that make up the QP will be described now, followed by the full QP formulation.

    \begin{figure*}
		\centering
		\begin{subfigure}[b]{0.31\textwidth}
			\centering
			\includegraphics[width=\textwidth]{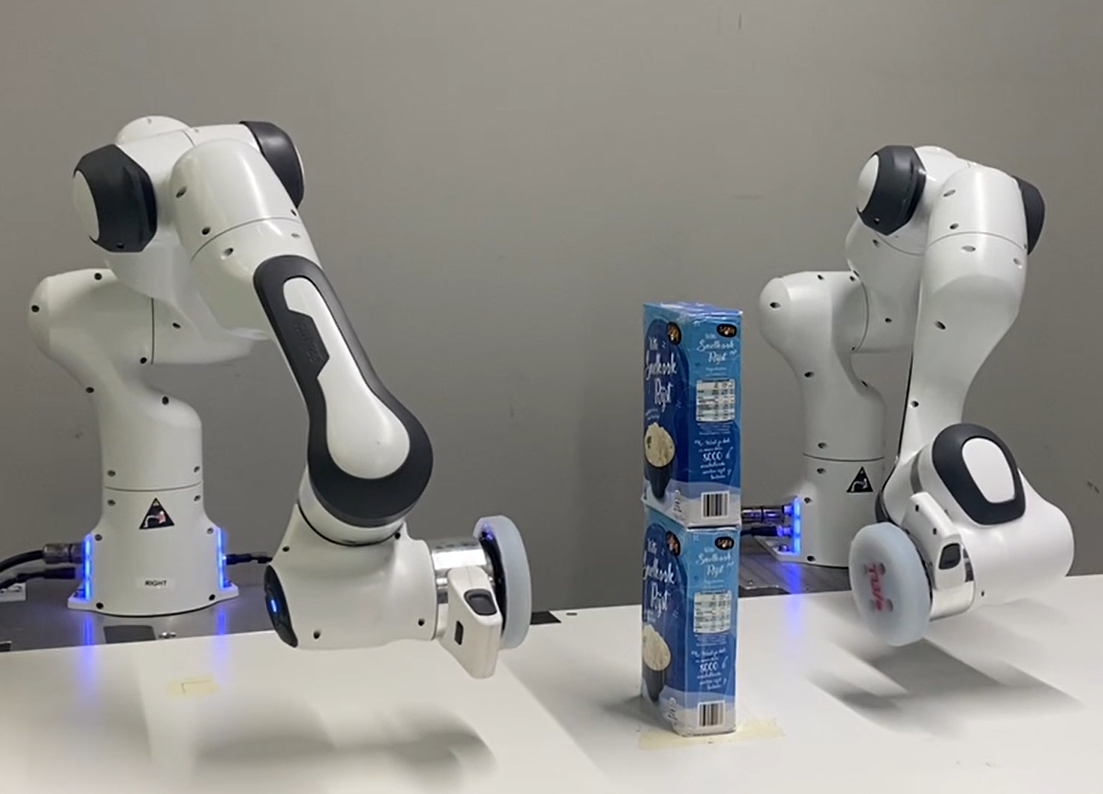}
			\caption{Ante-impact state}
			\label{fig:snap_ref_ante}
		\end{subfigure}
		\hfill
		\begin{subfigure}[b]{0.32\textwidth}
			\centering
			\includegraphics[width=\textwidth]{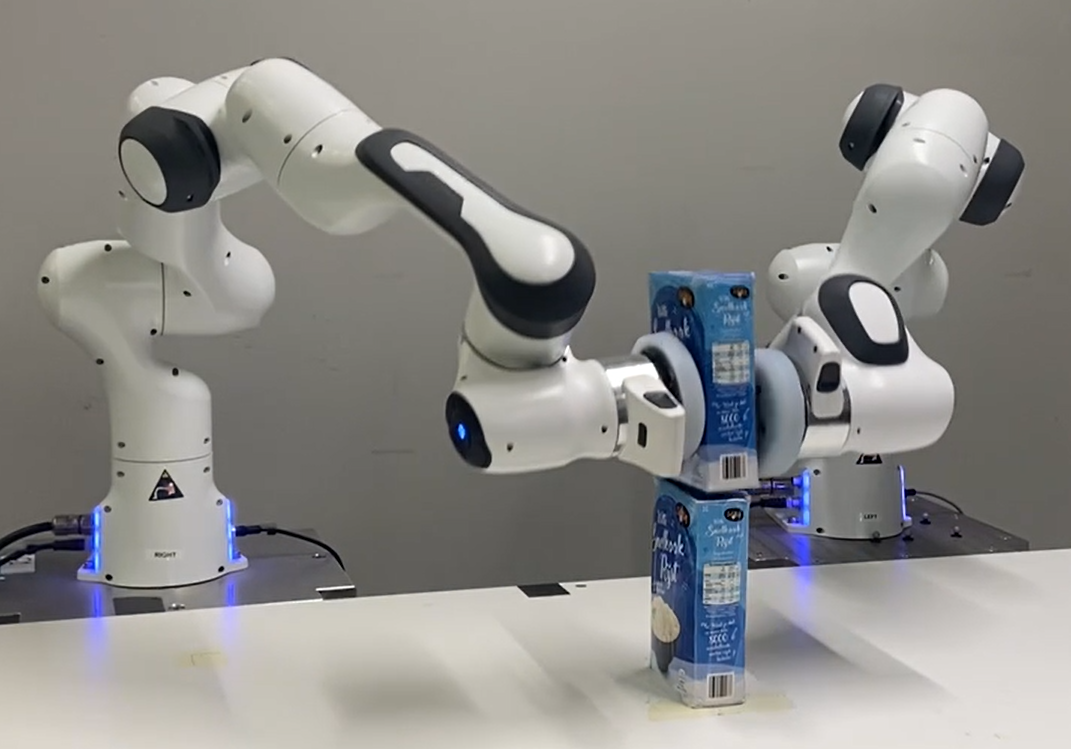}
			\caption{Simultaneous impact}
			\label{fig:snap_ref_interim}
		\end{subfigure}
		\hfill
		\begin{subfigure}[b]{0.31\textwidth}
			\centering
			\includegraphics[width=\textwidth]{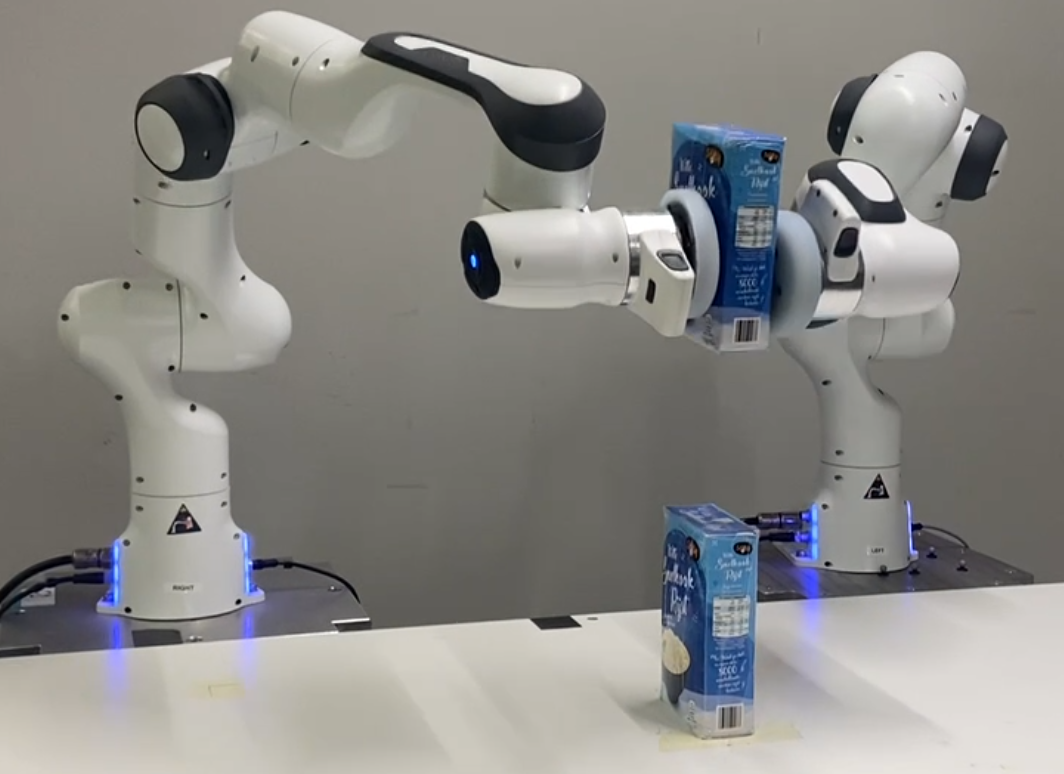}
			\caption{Post-impact state}
			\label{fig:snap_ref_post}
		\end{subfigure}
		\caption{Snapshots of the system during reference recording procedure.}
		\label{fig:snapshots_reference}
	\end{figure*}	
 
	\textbf{Impedance task:} Using the reference signals $\bm p_{i,d}$, $\bm R_{i,d}$ and $\vv_{i,d}$, the operator aims to control the end effector's movement and the wrench it exerts on the environment. This may be achieved with a Cartesian impedance controller \cite{Hogan1985}, which effectively makes the end effector behave as a mass-spring-damper system that is attached to the reference frame defined by $(\bm R_{i,d}, \bm p_{i,d})$. QP control  can be used to apply impedance control to kinematically redundant robots \cite{Hoffman2018} while controlling the null-space. We want to select input torques $\bm \tau_i$ in such a way that the apparent applied wrench at the end effector is identical to a desired spring-damper wrench $\vf_{i,r}$. We determine the force contribution of this wrench using a standard PD control law, while the torque contribution is inspired by the PD control law on SO(3) in 
	\cite{Bullo1995}, resulting in
	\begin{equation}\label{eq:imp_des}
	\vf_{i,r}= \bm D_{r} \left( \vv_{i,d}(t) - \vv_i \right) + \bm K_{r} \begin{bmatrix}\bm p_{i,d}(t) - {\bm p_{i}} \\ -(\log(\bm R_i^T{\bm R}_{i,d}(t)))^{\vee } \end{bmatrix}
	\end{equation}
	with operator $^\vee$ (vee) defined in accordance with \cite{Murray1994} as
	\begin{equation}\label{eq:hat3}
	\begin{bmatrix}
	0   & -z     &  y \\
	z   &  0 & -x     \\
	-y   &  x &  0 
	\end{bmatrix}^\vee  :=
	\begin{bmatrix} 
	x \\ 
	y \\
	z 
	\end{bmatrix}
	.
	\end{equation}
	The $6 \times 6$ matrix $\bm K_{r}$ is diagonal with user-defined entries, specifying the desired apparent stiffness between the end effector and its reference pose in each direction. Inspired by \cite{albu-schaffer2003}, we select for the damping gain 
	\begin{equation}\label{eq:D_imp}
	\bm D_{r} = \sqrt{\bm \Lambda_i} \sqrt{\bm K_{r}} + \sqrt{\bm K_{r}} \sqrt{\bm \Lambda_i}
	\end{equation}
	with task-space equivalent inertia matrix
	\begin{equation}
	\bm \Lambda_i := \left(\bm J_i\bm M_i^{-1} \bm J_{i}^T\right)^{-1}       
	\end{equation}
	to pursue a critically damped closed-loop response.  
	We then formulate a corresponding task error $\bm e_{i,\text{imp}}$ to be used in the cost function of the QP as 
	\begin{equation} \label{eq:e_imp_rec}
	\bm e_{i,r,\text{imp}} :=  \va_i - \bm \Lambda_i^{-1}\vf_{i,r}.
	\end{equation}
	If $\bm e_{i,r,\text{imp}}$ is brought to 0, which is pursued in the QP control scheme, the corresponding closed-loop behaviour of the end effector will indeed resemble that of an impedance control law, as is shown in Appendix A.
	
	\textbf{Posture task:} To address redundancy when tracking a 6DOF impedance task using 7DOF robots, an additional degree of freedom has to be prescribed to preserve uniqueness of the QP solution, while enforcing stable null-space behaviour. We also want to make sure that the null-space configuration at the time of impact is identical between teleoperation and autonomous control, as the impact dynamics are configuration-dependent \cite{Aouaj2021}. Without loss of generality, we chose to do so by formulating a so-called posture task that defines a desired acceleration $\beta_{i,r}$ for a given joint as 
	\begin{equation} \label{eq:beta_pos}  
	\beta_{i,r} := 2 \sqrt{k_{r,\text{pos}}}(\dot{\xi}_{i,d}(t) - \dot{\xi}_{i}) + k_{r,\text{pos}}({\xi}_{i,d}(t) - \xi_i),
	\end{equation}
	where the signal ${\xi}_i:=\bm S{\bm q}_i$ with selection matrix $\bm S \in \mathbb{R}^{1 \times 7}$ prescribes the position of a given joint, and $k_{r,\text{pos}}$ is a user-defined proportional gain. The index of the joint for which the task is prescribed, in our approach joint 1, was selected with maximum shared workspace and manipulability of the robots as criteria. A corresponding task error is then defined as
	\begin{equation} \label{eq:e_pos} e_{i,r,\text{pos}} :=  
	\ddot{\xi}_i - \beta_{i,r}.
	\end{equation}

	\textbf{Constraints:} To prevent the robots from exceeding any joint limits, constraints are defined that aim to prevent the robots from reaching the upper and lower limits of the joint positions, velocities and torques, defined as ${\bm q}_\text{max}, {\dot{\bm q}}_\text{max}$, ${\bm \tau}_\text{max}$ and ${\bm q}_\text{min}, {\dot{\bm q}}_\text{min}$, ${\bm \tau}_\text{min}$, respectively. Since the optimization variables are given by $\ddot{\bm q}_i$, the corresponding constraints are defined as in \cite{Bouyarmane2019} by
	\begin{equation}\label{eq:const_q}
	{\bm q}_\text{min} \leq \frac{1}{2}\ddot{\bm q}_i \Delta t^2 + \dot{\bm q}_i\Delta t + \bm q_i \leq {\bm q}_\text{max},
	\end{equation} 
	\begin{equation}\label{eq:const_dq}
	\dot{\bm q}_\text{min} \leq \ddot{\bm q}_i \Delta t + \dot{\bm q}_i \leq \dot{\bm q}_\text{max},
	\end{equation} 
	\begin{equation}\label{eq:const_tau}
	{\bm \tau}_\text{min} \leq \bm M_i\ddot{\bm q}_i + \bm h_i  \leq {\bm \tau}_\text{max},
	\end{equation}     
	with QP time step $\Delta t$. This implies that torque limit violation is prevented for the present time step, while position and velocity limits are prevented in a soft sense for the subsequent time step.

	\textbf{Full QP formulation:} Using the formulation of tasks and constraints given above, the full teleoperation-based impedance QP control approach is formulated as 
	\begin{equation}
	(\ddot{\bm q}_1^*, \ddot{\bm q}_2^*) = \underset{(\ddot{\bm q}_1, \ddot{\bm q}_2)}{\operatorname{argmin}} \ \sum_{i=1}^2 \left(w_{\text{imp}}\|{\bm e_{i,r,\text{imp}}}\|^2 + w_{\text{pos}} e^2_{i,r,\text{pos}} \right),
	\end{equation}
	subject to constraints \eqref{eq:const_q}-\eqref{eq:const_tau} for $i\in\{1,2\}$, with task weights $w_{\text{imp}}$ and $ w_{\text{pos}} \in \mathbb{R}^+$. As already mentioned, desired joint torques are then formulated using \eqref{eq:EOM_free}, which are sent to the robots.

    During demonstrations, the measured end effector 6D poses and velocities are recorded and stored as $\bm p_{i,r}(t)$, $\bm R_{i,r}(t)$, $\vv_{i,r}(t)$, as well as the joint position and velocity relevant to the posture task, stored as $\xi_{i,r}(t)$ and $\dot{\xi}_{i,r}(t)$. These signals will form the basis of the reference that is to be tracked autonomously using the RS-based control approach. We also record the desired wrench $\vf_{i,r}(t)$ from \eqref{eq:imp_des} and desired posture task acceleration $\beta_{i,r}(t)$ from \eqref{eq:beta_pos}, which will be used as basis for the feedforward reference. Note that especially saving $\vf_{i,r}(t)$ is essential for replicating contact forces, as these are not captured otherwise, which will become apparent in Section \ref{sec:post_mode}.  

	\subsection{Reference extension}\label{sec:reference_extension}

 
    In accordance with the core idea of RS to always have a reference at hand that corresponds to the current contact mode, as further explained in Section \ref{sec:introduction}, the recorded signals 
	are used to generate extended ante-impact and post-impact references signals that overlap around the nominal impact time. 
	This nominal impact time $T_r$ is extracted from the recording using an impact detection method, of which details are provided in Appendix B. However, the impact typically has a short but finite duration that is longer than that of a single time step. Furthermore, impact detection is typically slightly delayed, and directly after the impact, oscillatory responses are typically observed that are to be ignored in the reference extension procedure as these oscillations will die out over time. Therefore, a user-defined interval $\Delta T_r$ before and after the impact is excluded when splitting the recording. 
	Regarding the reference extension procedure, we first consider the feedforward signal $\vf_{i,r}(t)$, which is extended by means of a constant hold, resulting in, respectively, the extended ante-impact and post-impact signals $\bar{\vf}^a_{i,r}(t)$ and $\bar{\vf}^p_{i,r}(t)$ defined as
    \begin{align}
	\bar{\vf}^a_{i,r}(t) &=  {\vf}_{i,r}(\min(t, T_a)), \  \bar{\vf}^p_{i,r}(t) =  {\vf}_{i,r}(\max(t, T_p)), 
	\end{align}
	with $T_a := T_{r}-\Delta T_r$ and $T_p := T_{r}+\Delta T_r$. 
    The posture task feedforward signals and velocity signals are extended by means of a similar constant hold, resulting in $\bar{\beta}^a_{i,r}(t)$, $\bar{\vv}^a_{i,r}(t)$, $\bar{\dot{\xi}}^a_{i,r}(t)$, $\bar{\beta}^p_{i,r}(t)$, $\bar{\vv}^p_{i,r}(t)$ and $\bar{\dot{\xi}}^p_{i,r}(t)$. 
	To ensure consistency between the extended velocity and position references, we extend $\bm p_{i,r}(t)$ through integration of the corresponding velocity reference, resulting in extended ante- and post-impact position references $\bar{\bm p}^a_{i,r}(t)$ and $\bar{\bm p}^p_{i,r}(t)$ as
	\begin{equation}\label{eq:p_ante_ext}
	\bar{\bm p}_{i,r}^a(t) = 
	\begin{cases} 
	\bm p_{i,r}(t) & \text{if } t\leq T_a,\\
	\bm p_{i,r}(T_a) + \bm v_{i,r}(T_a)(t - T_a) & \text{if } t>T_a,
	\end{cases}
	\end{equation}
	\begin{equation}\label{eq:p_post_ext}
	\bar{\bm p}_{i,r}^p(t) = 
	\begin{cases} 
	\bm p_{i,r}(t) & \text{if } t\geq T_p,\\
	\bm p_{i,r}(T_p) + \bm v_{i,r}(T_p)(T_p -t) & \text{if } t<T_p,
	\end{cases}
	\end{equation}
	where the ante- and post-impact linear velocity reference are obtained from the respective twist reference as $\bar{\vv}^a_{i,r} = \left[\bar{\bm v}^a_{i,r}, \bar{\bm \omega}^a_{i,r}\right]$, $\bar{\vv}^p_{i,r} = \left[\bar{\bm v}^p_{i,r}, \bar{\bm \omega}^p_{i,r}\right]$. 
    In similar fashion to \eqref{eq:p_ante_ext} and \eqref{eq:p_post_ext}, we can extend $\xi_{i,r}(t)$ to the ante- and post-impact references $\bar{\xi}^a_{i,r}(t)$ and $\bar{\xi}^p_{i,r}(t)$ using the corresponding velocity references. 
    The extended ante- and post-impact orientation reference $\bar{\bm R}^a_{i,r}(t)$ and $\bar{\bm R}^p_{i,r}(t)$ are obtained from $\bm R_{i,r}(t)$ using integration of the corresponding angular velocity as
	\begin{equation}
	\bar{\bm R}_{i,r}^a(t) = 
	\begin{cases} 
	\bm R_{i,r}(t)& \text{if } t\leq T_a,\\
	\bm R_{i,r}(T_a)\exp\left(\bm \omega^\wedge_{i,r}(T_a)(t-T_a)\right)& \text{if } t> T_a,
	\end{cases}
	\end{equation}
	\begin{equation}
	\bar{\bm R}_{i,r}^p(t) = 
	\begin{cases} 
	\bm R_{i,r}(t)& \text{if } t \geq T_p,\\
	\bm R_{i,r}(T_p)\exp\left({\bm \omega}^\wedge_{i,r}(T_p)(T_p - t)\right)& \text{if } t < T_p
	\end{cases}
	\end{equation}
	with operator $^\wedge$ (hat) in accordance with \cite{Murray1994} defined as
	\begin{equation}\label{eq:vee3}
	\begin{bmatrix} 
	x \\ 
	y \\
	z 
	\end{bmatrix}^{\wedge}
	:=
	\begin{bmatrix}
	0   & -z     &  y \\
	z   &  0 & -x     \\
	-y   &  x &  0 
	\end{bmatrix}.
	\end{equation}

    \begin{figure}		
		\centering
		\includegraphics[width=\linewidth]{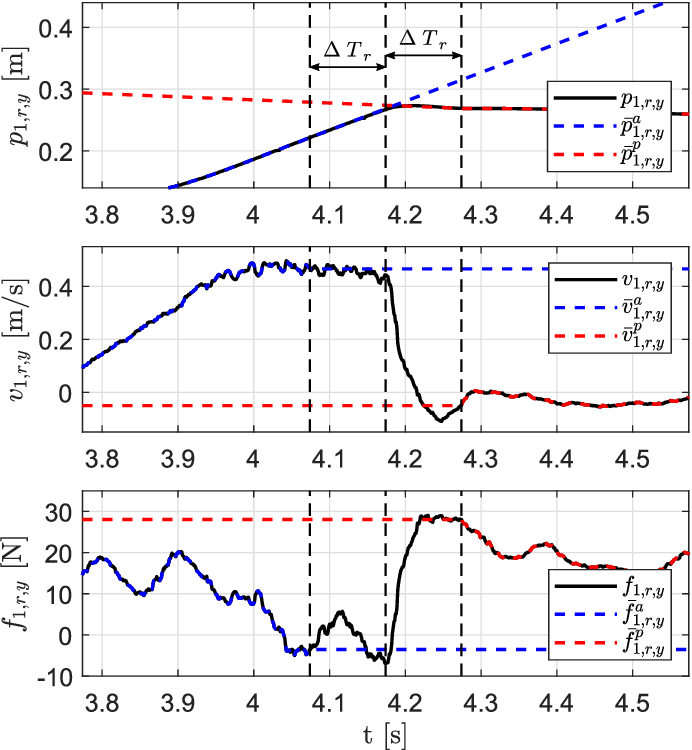}
		\caption{Depiction of the recorded position, velocity and force signals around the impact time in $y$-direction, i.e. the direction normal to the impact surface, together with the extended ante- and post-impact references. The black dashed lines indicates the recorded impact time $T_r$ together with the time interval of exclusion $\Delta T_r$.}
		\label{fig:reference}
	\end{figure}
 
	Snapshots of the system during reference recording via teleoperation for the use case considered in this paper are depicted in Figure \ref{fig:snapshots_reference}. The position, velocity and desired wrench of the end effector of robot 1 in the $y$-direction normal to the impact surface obtained from this recording are shown in Figure \ref{fig:reference}, together with the extended ante- and post-impact reference extracted from these signals. Depending on the contact state of the manipulator, the corresponding reference will be used for feedback control, as will be explained in the Section \ref{sec:control_approach}. 


	\section{Reference spreading based QP control approach}\label{sec:control_approach}
	
	In Section \ref{sec:reference_generation}, a method to acquire a reference motion based on QP-based robot control and teleoperation is detailed together with a concrete example, that is provided for illustration. This motion is then used to generate extended ante- and post-impact references that match the impact dynamics of the system, which will now be used to construct a RS-based control action without a human in the loop. In this section, we detail how ante-impact, interim, and post-impact control actions can be realized via suitable QPs, obtaining for each mode a suitable joint torque $\bm \tau_i^*$, with $i \in \{1,2\}$, to be applied to the robot arms after every time step $\Delta t$.


	\subsection{Ante-impact mode}
	
	\begin{figure*}[t]
		\centering
		\includegraphics[trim={1cm 17.0cm 5.3cm 7.2cm}, clip]{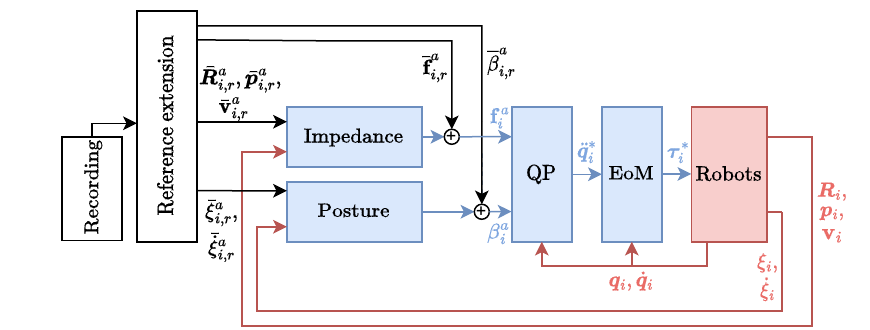}
		\caption{Visualization of the ante-impact QP control scheme employed for tracking the recorded reference using the proposed RS-based control framework. The  physical robot signals are red, the controller-related signals are blue, and extended reference signals are black.}
		\label{fig:flow_rep}
	\end{figure*}
	
	The ante-impact mode is initially active until a first impact is detected using the approach detailed in Appendix B. Its objective is to establish near-simultaneous contact in the desired ante-impact state provided by the corresponding ante-impact reference. 
	The ante-impact QP controller is very similar in structure to the QP controller defined in Section \ref{sec:tele_impedance} used to generate the reference signals. 
	The main difference is that the impedance and posture task are set to follow the extended ante-impact reference signals defined in Section \ref{sec:reference_extension}, with additional feedforward control. Furthermore, the control gains are considerably higher to enable accurate autonomous tracking, as discussed in Section \ref{sec:introduction}. A similar diagram to Figure \ref{fig:flow_rec} corresponding to the ante-impact mode is shown in Figure \ref{fig:flow_rep} to give a visual overview of the ante-impact QP controller, with further details provided below.

    First, the impedance task is modified by defining a desired ante-impact wrench $\vf^a_i$ in similar fashion to \eqref{eq:imp_des} as
	\begin{equation}\label{eq:imp_des_ante}
	\begin{aligned}
	\vf^a_{i} := & \ \bar{\vf}^a_{i,r}(t)  + \bm D^a \left( \bar{\vv}^a_{i,r}(t) - \vv_i \right) \\ & + \bm K^a \begin{bmatrix}\bar{\bm p}^a_{i,r}(t) - {\bm p_{i}} \\ -(\log(\bm R_i^T\bar{\bm R}^a_{i,r}(t)))^{\vee } \end{bmatrix}.
	\end{aligned}
	\end{equation}
	with user-defined $\bm K^a$, and $\bm D^a$ selected accordingly as in \eqref{eq:D_imp}. We then define the impedance task error $\bm e_{i,\text{imp}}$ in similar fashion to \eqref{eq:e_imp_rec} as
	\begin{equation} \label{eq:e_imp_rec_ante}
	\bm e_{i,\text{imp}} =  \va_i - \bm \Lambda_i^{-1}\vf^a_{i}.
	\end{equation}
	The posture task is also updated as the new desired joint acceleration $\beta_i^a$ is redefined based on \eqref{eq:beta_pos} as 
	\begin{equation} \label{eq:beta_pos_ante}  
	\beta^a_{i} = \bar{\beta}^a_{i,r}(t) + 2 \sqrt{k^a_{\text{pos}}}(\bar{\dot{\xi}}^a_{i,r}(t) - \dot{\xi}_{i}) + k^a_{\text{pos}}(\bar{\xi}^a_{i,r}(t) - \xi_i)
	\end{equation}
	with user-defined proportional gain $k^a_{\text{pos}}$. We then re-define a corresponding task error $e_{i,\text{pos}}$ as     
	\begin{equation} \label{eq:e_pos_ante} e_{i,\text{pos}} =  
	\ddot{\xi}_i - \beta^a_{i}.
	\end{equation}
	This leads to the full QP controller 
	\begin{equation}\label{eq:QP}
	(\ddot{\bm q}_1^*, \ddot{\bm q}_2^*) = \underset{\ddot{\bm q}_1, \ddot{\bm q}_2}{\operatorname{argmin}} \ \sum_{i=1}^2 \left(w_{\text{imp}}\|{\bm e_{i,\text{imp}}}\|^2 + w_{\text{pos}} \left(e_{i,\text{pos}}\right)^2 \right),
	\end{equation}
	subject to constraints \eqref{eq:const_q}-\eqref{eq:const_tau} for $i\in\{1,2\}$. Desired joint torques are again formulated using \eqref{eq:EOM_free}.

	\subsection{Interim mode}

    As mentioned in Section \ref{sec:introduction}, the interim mode is active after the first impact is detected, since uncertainty can cause contact to be only partially established at this time. 
	In the following, we propose a corresponding interim mode controller that is active for a pre-defined time duration $\Delta t_\text{int}$, starting from the moment the first impact is detected through the impact detection approach highlighted in Appendix B, called $T_\text{imp}$. We select $\Delta t_\text{int}$ conservatively in such a way that, despite some acceptable level of uncertainties, full contact will be established within this time based on recorded experiments. Furthermore, we assume sufficient dissipation in the interim mode is provided by the environment, implying that active damping to reduce the effect of post-impact vibrations, as e.g. proposed in \cite{Khatib1986}, is not required.
	
	In line with previous works on RS with simultaneous impacts \cite{Steen2022, Steen2022a, Steen2022b}, we want to initially disable velocity tracking control in the interim mode given the lack of a velocity reference that matches the uncertain contact state, and apply feedforward and position feedback based on the ante-impact references to ensure input continuity with the ante-impact mode. However, as we opt to select a conservatively large value for $\Delta t_\text{int}$, it is possible that the robot state would diverge significantly from the desired post-impact robot-state if we apply control based on the ante-impact reference for the entire duration of the interim mode as done in \cite{Steen2022}. Such an interim mode design could lead to a large jump in the velocity tracking error and large resulting input steps when transitioning to the post-impact mode. Hence, it is opted to define the interim mode desired wrench $\vf^\text{int}_i$ as
	\begin{equation}\label{eq:imp_des_int}
	\begin{aligned}
	\vf^\text{int}_{i} := & \ (1-\gamma)\bar{\vf}^a_{i,r}(t) +  \gamma\bar{\vf}^p_{i,r}(t) \\ + & \ \bm D^\text{int}_{i} \left( (1-\gamma)\vv_i + \gamma \bar{\vv}^p_{i,r}(t) - \vv_i \right) \\  + & \ \bm K^\text{int} \begin{bmatrix} (1-\gamma)\bar{\bm p}^a_{i,r}(t) +  \gamma\bar{\bm p}^p_{i,r}(t) - {\bm p_{i}} \\ -(\log(\bm R_i^T \bm R^\text{int}_{i,d}))^{\vee } \end{bmatrix}
	\end{aligned}
	\end{equation}
	with user-defined $\bm K^\text{int}$, with $\bm D^\text{int}$ again selected as in \eqref{eq:D_imp}, with
	\begin{equation}
	\bm R^\text{int}_{i,d} := \bar{\bm R}^a_{i,r}(t) \exp\left(\gamma \log\left([\bar{\bm R}^a_{i,r}(t)]^T\bar{\bm R}^p_{i,r}(t)\right)\right)
	\end{equation}
	and with so-called blending parameter $\gamma$ defined as
	\begin{equation}
	\gamma := \frac{t-T_\text{imp}}{\Delta t_\text{int}}.
	\end{equation}
	Please note that the ante-impact velocity reference $\bar{\vv}^a_{i,r}(t)$ that one would expect to see in \eqref{eq:imp_des_int} is instead replaced by the measured velocity $\vv_i$. At the start of the interim mode, i.e. when $\gamma = 0$, this implies the velocity feedback term in \eqref{eq:imp_des_int} is equal to zero, removing velocity feedback at the beginning of the interim mode to avoid peaking during the uncertain contact state. The feedforward and position feedback terms of \eqref{eq:imp_des_int} for $\gamma = 0$ are, however, identical to those in \eqref{eq:imp_des_ante}. 
    Meanwhile, at the end of the interim mode, the desired wrench is identical to the desired post-impact wrench $\vf_i^p$ as will become apparent later in this section. This implies that the transition from the ante-impact to the interim mode and from the interim to the post-impact mode are smooth, hence removing any possible input step at switching times that would result when using previously formulated interim mode formulations \cite{Steen2022,Steen2022a,Steen2022b}. Meanwhile, the overall reduction of velocity feedback, especially early in the interim phase, implies that the effect of rapid and unpredictable velocity jumps on the input torques is greatly reduced, while feedforward and position feedback using the ante-impact references is used during this time to make sure the impact event is successfully finished.

	
	Using \eqref{eq:imp_des_int}, we can redefine the impedance task error during the interim mode as
	\begin{equation} \label{eq:e_imp_rec_int}
	\bm e_{i,\text{imp}} =  \va_i - \bm \Lambda_i^{-1}\vf^\text{int}_{i}.
	\end{equation}
	In a similar fashion, we redefine the posture task to obtain desired joint acceleration $\beta^\text{int}_{i}$ as
	\begin{equation} \label{eq:beta_pos_int}  
	\begin{aligned}
	\beta^\text{int}_{i} := & \ (1 - \gamma) \bar{\beta}^a_{i,r}(t) + \gamma \bar{\beta}^p_{i,r}(t) 
    \\ +& \  2 \sqrt{k^\text{int}_{\text{pos}}}((1-\gamma)\dot{\xi}_{i} +
     \gamma\bar{\dot{\xi}}^p_{i,r}(t) - \dot{\xi}_{i}) \\ + & \  k^\text{int}_{\text{pos}}((1 - \gamma) \bar{\xi}^a_{i,r}(t) + \gamma \bar{\xi}^p_{i,r}(t) - \xi_i)
	\end{aligned}
	\end{equation}
	with user-defined proportional gain $k^\text{int}_{\text{pos}}$ and task error     
	\begin{equation} \label{eq:e_pos_int} e_{i,\text{pos}} =  
	\ddot{\xi}_i - \beta^\text{int}_{i}.
	\end{equation}
	The full interim mode QP controller is then given by \eqref{eq:QP}, using $e_{i,\text{imp}}$ and $e_{i,\text{pos}}$ as formulated in \eqref{eq:e_imp_rec_int} and \eqref{eq:e_pos_int}, respectively. 

	\subsection{Post-impact mode}\label{sec:post_mode}
	
	The post-impact control mode is entered at time $T_\text{imp} + \Delta t_\text{int}$ as mentioned earlier in this section. The objective in the post-impact mode in dual-arm manipulation is to not only follow a desired reference pose with the end effectors, but also maintain sufficient force to manipulate the object. Despite this different objective, the controller is almost identical to the ante-impact mode, only replacing the ante-impact reference signals with the post-impact reference signals. 
	The reason why the post-impact mode does not require any structural update compared to the ante-impact mode is that the desired contact force is maintained through the feedforward signal $\bar{\vf}^p_{i,r}(t)$. This signal contains both the wrench used to accelerate the robots and the object, as well as the wrench required to clamp the object with a desired force.
 
	To summarize, by using a reference associated to the current contact state in the ante-impact and post-impact mode, and by blending between the ante-impact and post-impact references with reduced velocity feedback in the interim mode, the peaking phenomenon, mentioned in Section \ref{sec:introduction}, is avoided. This implies that impact-induced input peaks and steps are avoided throughout all of the three control modes.

	\section{Experimental validation}\label{sec:validation}
 
\begin{figure*}
		\centering
		\begin{subfigure}[b]{0.315\linewidth}
			\centering
			\includegraphics[width=\textwidth]{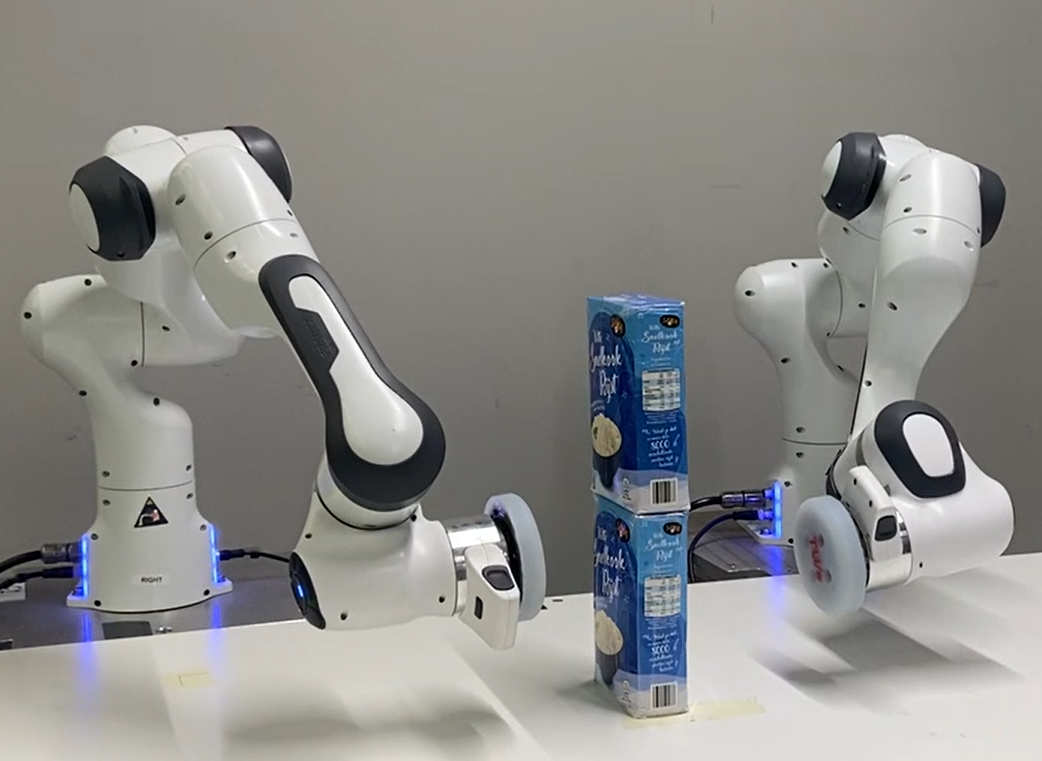}
			\caption{Ante-impact state}
			\label{fig:snap_track_ante}
		\end{subfigure}
		\hfill
		\begin{subfigure}[b]{0.335\linewidth}
			\centering
			\includegraphics[width=\textwidth]{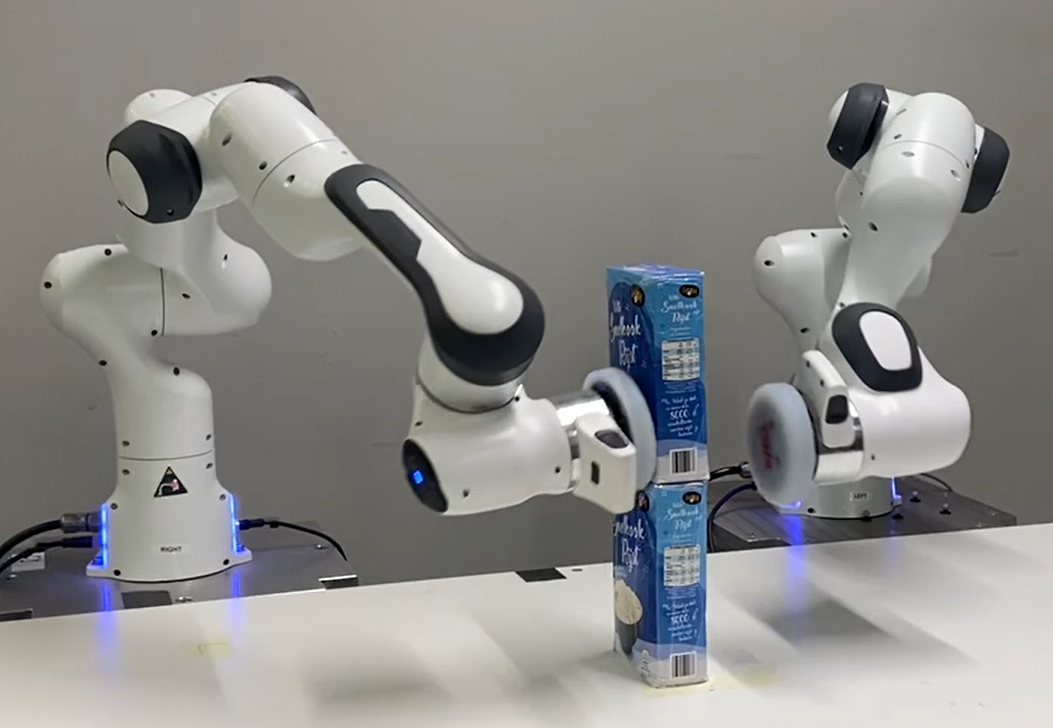}
			\caption{Non-simultaneous impact}
			\label{fig:snap_track_interim}
		\end{subfigure}
		\hfill
		\begin{subfigure}[b]{0.305\linewidth}
			\centering
			\includegraphics[width=\textwidth]{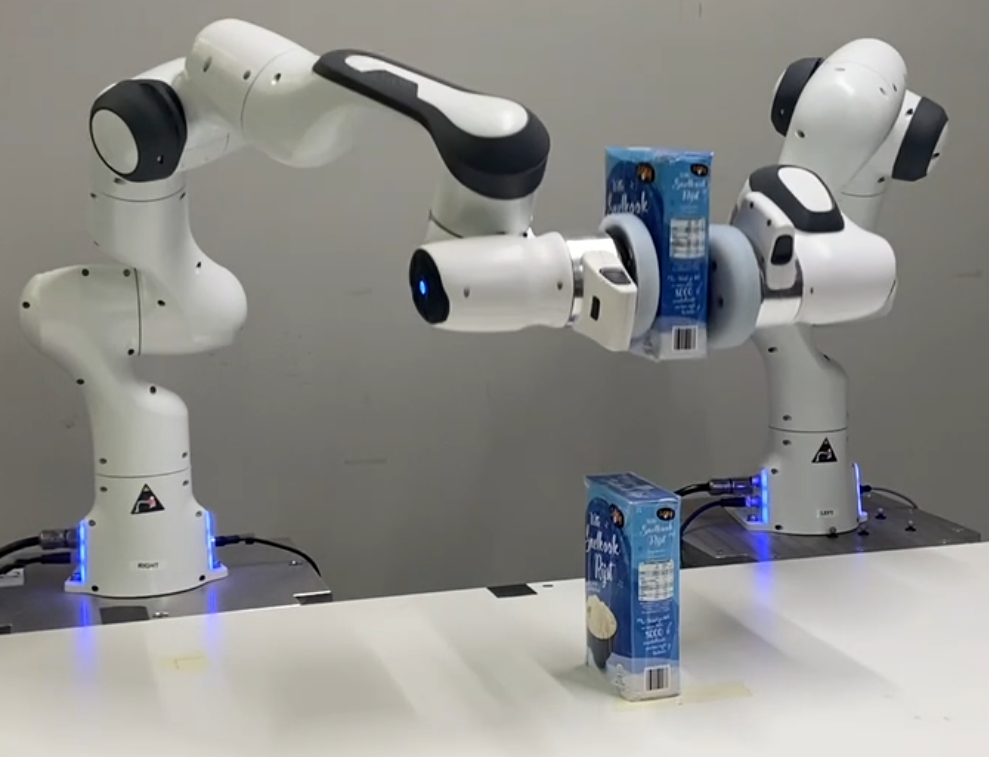}
			\caption{Post-impact state}
			\label{fig:snap_track_post}
		\end{subfigure}
		\caption{Snapshots of the system during autonomous control. By adding an offset in the object location compared to the reference recording procedure, environmental uncertainty is simulated, causing a non-simultaneous impact.}
		\label{fig:snapshots_tracking}
	\end{figure*}

	In order to experimentally validate\footnote{A video showcasing the experimental validation can be found here: \href{https://youtu.be/jKzKd-z-EGE}{https://youtu.be/jKzKd-z-EGE}.} the proposed framework, the control approach described in Section \ref{sec:control_approach} is used to autonomously perform a grabbing task for a variety of objects using multiple references generated through the approach described in Section \ref{sec:reference_extension}. We have used the QP solver provided by the software framework mc\_rtc\footnote{mc\_rtc: \href{https://jrl-umi3218.github.io/mc\_rtc/}{https://jrl-umi3218.github.io/mc\_rtc/}} to perform these experiments. To emulate uncertainty in the environment, the to-be-manipulated object is shifted in the y-direction of the world frame defined in Figure \ref{fig:setup} over a range of distances during different experiments. As depicted in Figure \ref{fig:snapshots_tracking}, which shows snapshots of an experiment where the object displacement was -30 mm, this can result in a non-simultaneous impact, where one of the robots impacts the object before the other one, and before the nominal impact time $T_r$. 

    \begin{table}[]
    \caption{Table containing parameter values used in reference recording procedure and proposed control approach.}
\begin{tabular}{l|l}
Parameter                                                                                       & Value                                  \\ \hline
$\Delta t$                                                               & 0.001                                    \\
$\Delta T_r$, $\Delta t_\text{int}$                                                               & 0.1                                    \\
$K_{r}$                                                                                         & $\text{diag}(300,300,300,20,20,20)$    \\
$\bm K^a$, $\bm K^\text{int}$, $\bm K^p$                                                                    & $\text{diag}(2000,2000,2000,20,20,20)$ \\
$k_{r,\text{pos}}$, $k^a_{\text{pos}}$, $k^\text{int}_{\text{pos}}$, $k^p_{\text{pos}}$         & 500                                    \\
$w_{\text{imp}}$,$w_{\text{pos}}$ & 1
\end{tabular}
\label{tab:parameters}
\end{table}

    The proposed approach is compared against three baseline approaches. In the first baseline approach, referred to as the approach with \textit{no RS}, we assume the ante- and post-impact reference do not overlap around the impact time, implying that we switch from the ante-impact mode directly to the post-impact mode at the nominal impact time $T_r$ rather than at the detected impact time $T_\text{imp}$. 

    In the second baseline approach, referred to as the approach with \textit{no velocity feedback}, we still switch from the ante-impact to the post-impact mode at the nominal impact time, but we additionally remove velocity feedback entirely around the time of the nominal impact event to reduce the effect of peaking. This approach is similar to that of \cite{Yang2021}, which removes velocity feedback in the subspace affected by the impact for a pre-defined time range. We select this time range as $\Delta t_\text{int}$ seconds before and after the nominal impact event, recalling that $\Delta t_\text{int}$ is the pre-defined duration of the interim mode in the proposed approach. The amount of time without velocity feedback is thus twice as much as that of the interim mode, which is chosen as the interim mode only needs to cover the duration of the impact sequence, while the velocity feedback removal additionally has to cover the uncertainty of the initial impact timing.

    In the third baseline approach, referred to as the approach with \textit{no interim mode}, we do take into account the actual impact time $T_\text{imp}$. However, rather than switching to the interim mode, we switch directly to the post-impact mode, as done in previous iterations for RS with \textit{single} impacts \cite{Saccon2014, Rijnen2020}.

	\begin{figure*}
		\centering
		\includegraphics[width=\linewidth,clip]{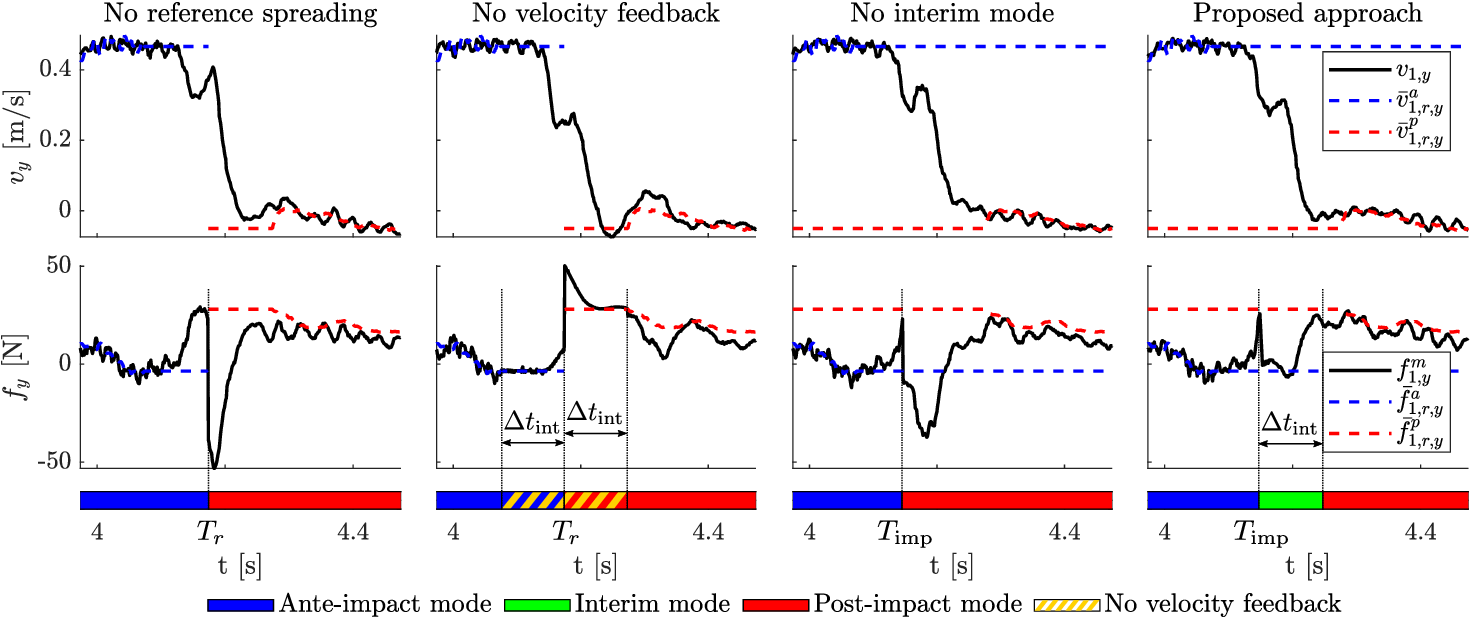}
		\caption{Velocity and desired wrench in $y$-direction of the end effector of robot 1 during a representative experiment with the rice box object, compared against the ante- and post-impact velocity reference and the ante- and post-impact feedforward wrench. Black dotted lines indicate mode transitions, with nominal impact time $T_r$, recorded impact time $T_\text{imp}$, and interim mode duration $\Delta t_\text{int}$.}
		\label{fig:results}
	\end{figure*}

    The parameter values used for all of the performed experiments are shown in Table \ref{tab:parameters}. Please note that the apparent stiffness matrices $\bm K^a$, $\bm K^\text{int}$ and $\bm K^p$ in the different modes used for the impedance task are identical, allowing for a fair comparison between the baselines and the proposed approach. While a summary of the results for all experiments is shown later in this section, we first focus on one of these experiments for the grabbing of a 1.25 kg rice box with an object displacement of -30 mm. This experiment, which is deemed representative, is used to highlight the beneficial effects of the proposed approach.   
    The resulting end effector velocities and desired wrenches of robot 1 in $y$-direction normal to the impacting surfaces for the representative experiment are reported in Figure \ref{fig:results}, both for the three baseline approaches as well as for the proposed approach. As predicted given the displacement of the object, an initial impact between robot 1 and the object occurs before sustained contact is eventually established with both end effectors, apparent from two distinct velocity jumps for each of the four control approaches. 
	
	Considering the desired force in the approach with no RS in the left of Figure \ref{fig:results}, it can be seen that after the initial impact, marked by the first velocity jump, a positive force is commanded, imposing an acceleration of the end effector towards the object. This is explained by the fact that the robot is trying to follow the ante-impact reference, which commands a higher velocity than that of the end effector after the first impact. Due to this acceleration, the velocity at the nominal impact time is significantly higher than the post-impact velocity reference, which is followed after the nominal impact time. This causes the velocity tracking error to spike up, resulting in a significant and undesired step in the desired force, as shown in the figure. 

    In the approach with no velocity feedback, second from the left in Figure \ref{fig:results}, a small peak in the desired force can be observed at the moment of switching from the ante-impact to the post-impact mode. This peak mainly results from the discontinuity between the ante- and post-impact feedforward force, but is small compared to the peak shown for the approach with no RS. However, with the lack of any velocity feedback, a big vibration can be observed in the velocity signal after the impact, resulting in a temporary loss of contact between the robot and the object. This temporary loss of contact can result in a poor grasp or even dropping of the object, which is definitely undesired. 
 
    In the approach with no interim mode, shown as third from the left in Figure \ref{fig:results}, a less extreme, but similar undesired effort can be seen as in the approach with no RS. Here, the system switches directly to the post-impact mode at the moment of impact detection before full contact is established. This also results in a large velocity tracking error, causing an unnecessary negative desired force and a resulting deceleration.
	
    To the contrary, the desired force only shows a minor peak in the proposed approach, depicted on the right in Figure \ref{fig:results}, which results from a slightly delayed detection of the first impact. As soon as the impact is detected, the system enters the interim mode, and the desired force jumps back to the level before the impact occurred. This is in accordance with the philosophy of the interim mode, which is identical to the ante-impact mode at the beginning of the interim mode,  while only removing velocity error feedback that is unreliable due to the uncertain contact state. The desired force then gradually approaches that of the post-impact mode as a result of the blending of the ante-and post-impact references. The continuity in the desired force during the transition from interim to post-impact mode is also in accordance with the philosophy of the proposed approach, which ensures the QP task errors in the interim and post-impact mode are identical at the moment of transitioning. 
    Despite the initial removal of velocity feedback when entering the interim mode, the large vibration seen in the approach with velocity feedback is avoided by gradually increasing the velocity error feedback.  
    
    In summary, only in the proposed approach is the risk of destabilization reduced as input peaks and steps are avoided, while a big vibration upon contact establishment is also avoided. Despite not being shown here, a similar reduction of the force peak is observed for robot 2 when using the proposed approach compared to the baseline approach with no RS, and with no interim mode, as well as a reduction of the effort related to the posture task. And, also for robot 2, a vibration upon contact establishment is observed only for the approach with no velocity feedback.

    To further validate the approach, a total of ten different references have been generated to grasp the 1.25 kg rice box using the teleoperation-based approach described in Section \ref{sec:reference_generation}, resulting in a natural variety of grasping motions. Initial experiments using the proposed approach and the three baseline approaches have then been performed for each reference, with a y-displacement of -30 mm and 30 mm, resulting in 80 total experiments. As a result of this displacement, not every experiment results in a successful grasp. Vibrations during contact establishment can result in the object being dropped, or safety limits of the Franka Emika robot\footnote{For more information on the safety limits of the Franka Emika robot, see: \href{https://frankaemika.github.io/docs/control\_parameters.html\#limits-for-panda}{https://frankaemika.github.io/docs/control\_parameters.html\#limits-for-panda}} being exceeded. The success rates for each control approach are depicted in the first column of Table \ref{tab:comparison}, where a successful grasp is defined as a grasp where the robot establishes and maintains sustained contact without triggering any safety limit before the prescribed release of the object. 
    
    \begin{table}[]
    \caption{Table containing 1) success rates for grab motions of a rice box using the proposed control approach and the three baseline approaches, using 10 different references and an object displacement of 30 and -30 mm, and 2) average difference for proposed and baseline approaches between maximum lifting height of the rice box and average maximum lifting height for experiments with identical reference and object displacement (as a measure for a successful grabbing action).}
    \centering
    \begin{tabular}{c|c|c}
    \multicolumn{1}{l|}{}  & \begin{tabular}[c]{@{}c@{}}Succesful grab\\ percentage {[}\%{]}\end{tabular} & \begin{tabular}[c]{@{}c@{}}Average max lifting\\ height difference {[}mm{]}\end{tabular} \\ \hline
    No reference spreading & 80                                                                           & 9.81                                                                                 \\ \hline
    No velocity feedback         & 40                                                                           & -17.07                                                                               \\ \hline
    No interim mode        & 85                                                                           & 0.78                                                                                 \\ \hline
    Proposed approach      & 90                                                                           & 6.47                                                                                
    \end{tabular}
    \label{tab:comparison}
    \end{table}

    From this table, it can be observed that the approach with no velocity feedback results in the the worst performance by far. This can be partially attributed to the aforementioned large vibration upon contact establishment also observed in Figure \ref{fig:results}, causing a delay in the establishment of sustained contact while both robots are moving up, resulting in the object to be dropped. But aside from this, the lack of damping in the approach also causes a significant increase in the percentage of experiments with safety limit violations upon impact, either in the maximum joint torque or the maximum joint velocity. This is clearly highly undesired. 
    The number of successful experiments significantly increases for the other two baseline approaches (no RS and no interim mode), but we still see that the proposed approach results in the largest number of successful experiments. This can be explained by the removal of the input peaks in the proposed approach, causing a reduction of the peak torques and hence reducing the number of times the safety limit for the maximum torque is violated. 

 \begin{figure*}
		\centering
		\begin{subfigure}[b]{0.1732\textwidth}
			\centering
			\includegraphics[width=\textwidth]{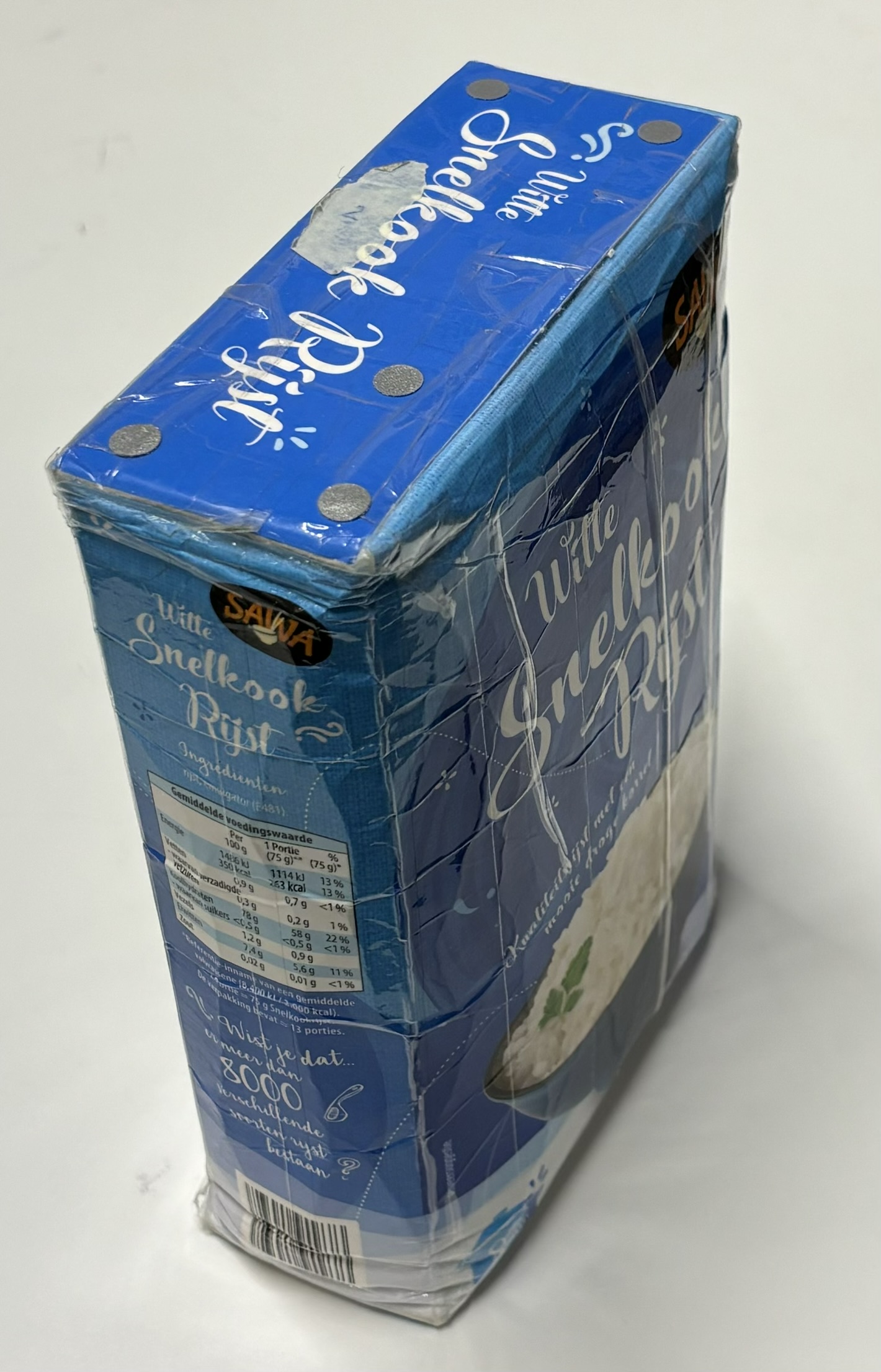}
			\caption{Rice box (1.25 kg)}
			\label{fig:snap_ref_ante}
		\end{subfigure}
        \hfill
  	\begin{subfigure}[b]{0.23\textwidth}
			\centering
			\includegraphics[width=\textwidth]{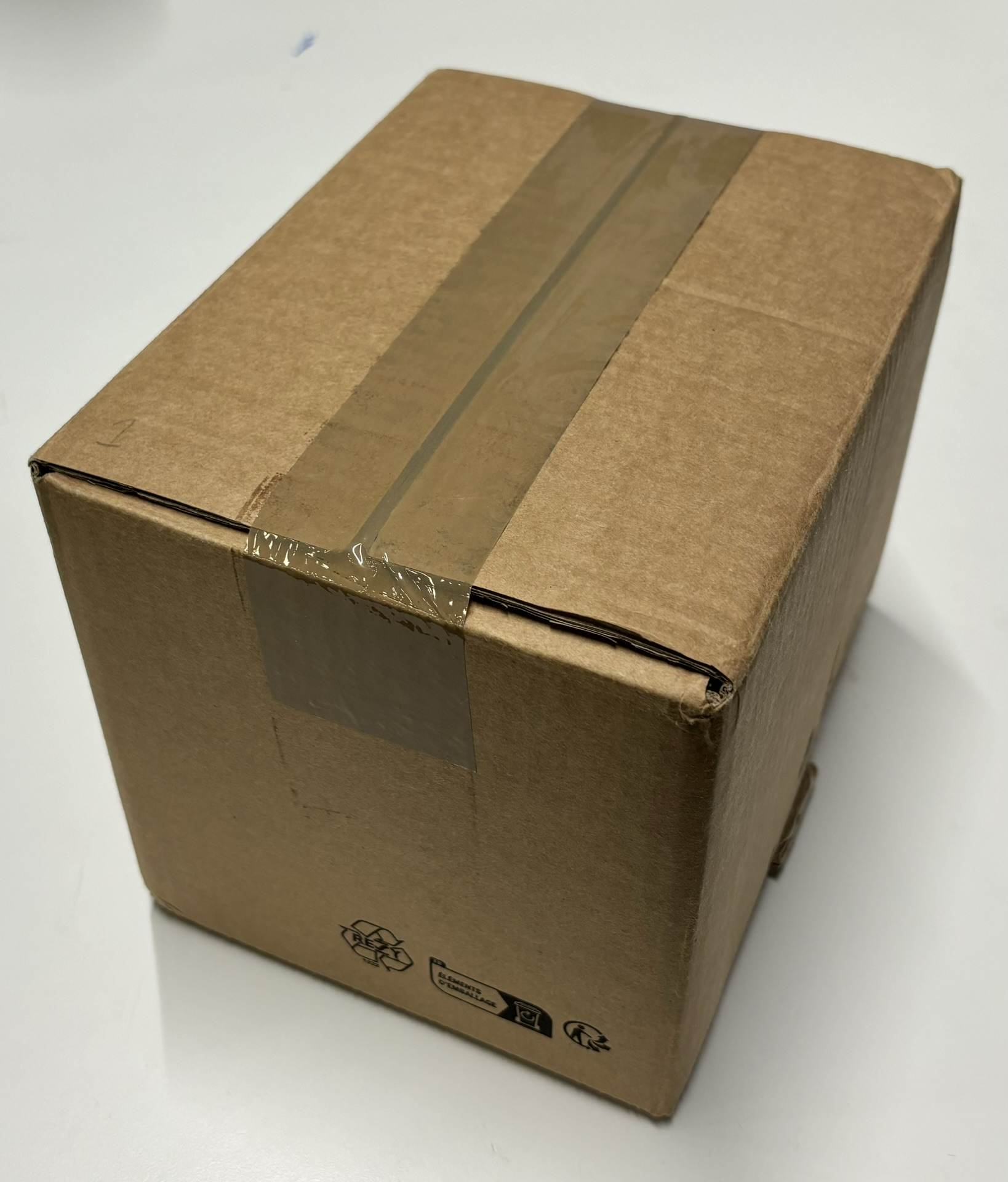}
			\caption{Single parcel (0.64 kg)}
			\label{fig:snap_ref_ante}
		\end{subfigure}
		\hfill
		\begin{subfigure}[b]{0.35\textwidth}
			\centering
			\includegraphics[width=\textwidth]{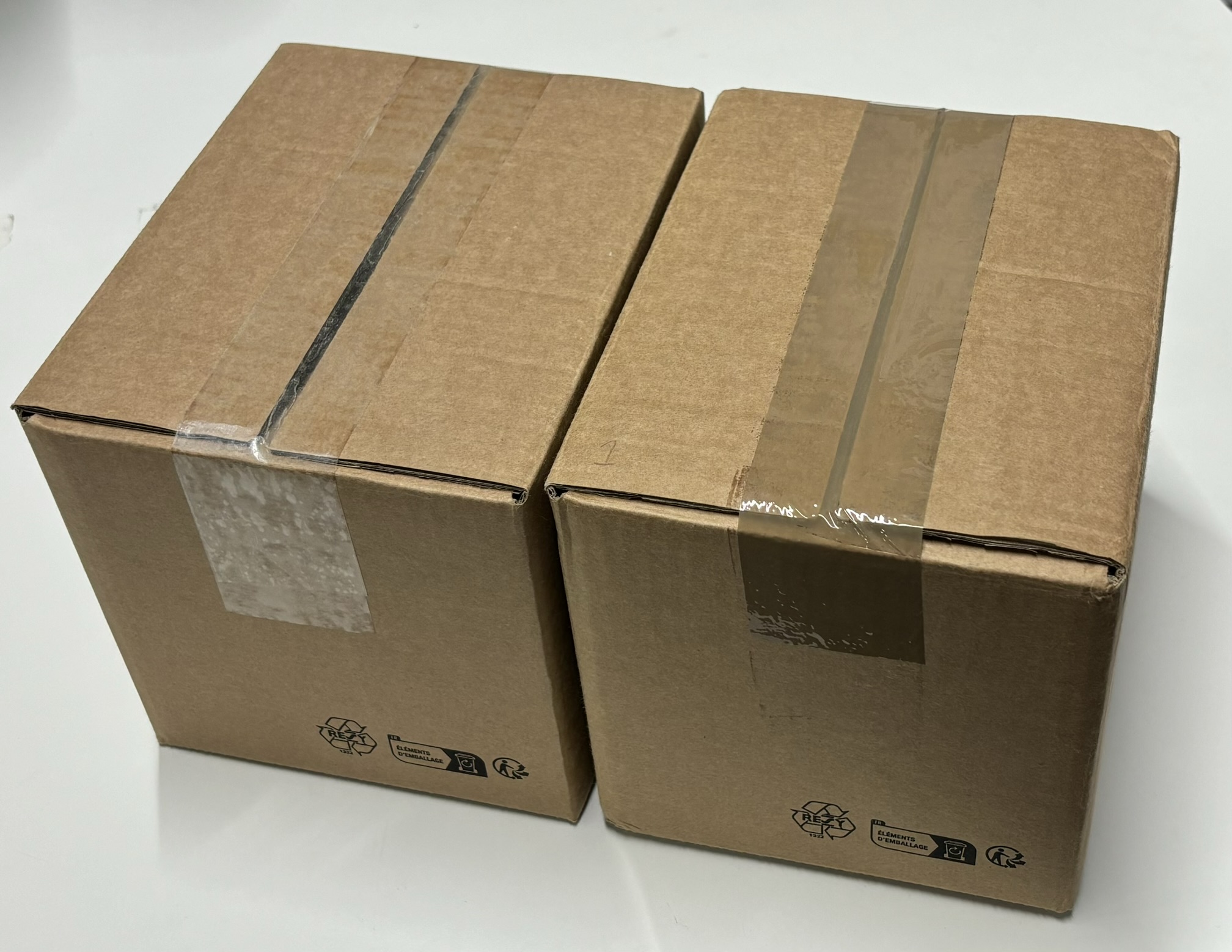}
			\caption{Two parcels (1.28 kg)}
			\label{fig:snap_ref_interim}
		\end{subfigure}
		\hfill
		\begin{subfigure}[b]{0.217\textwidth}
			\centering
			\includegraphics[width=\textwidth]{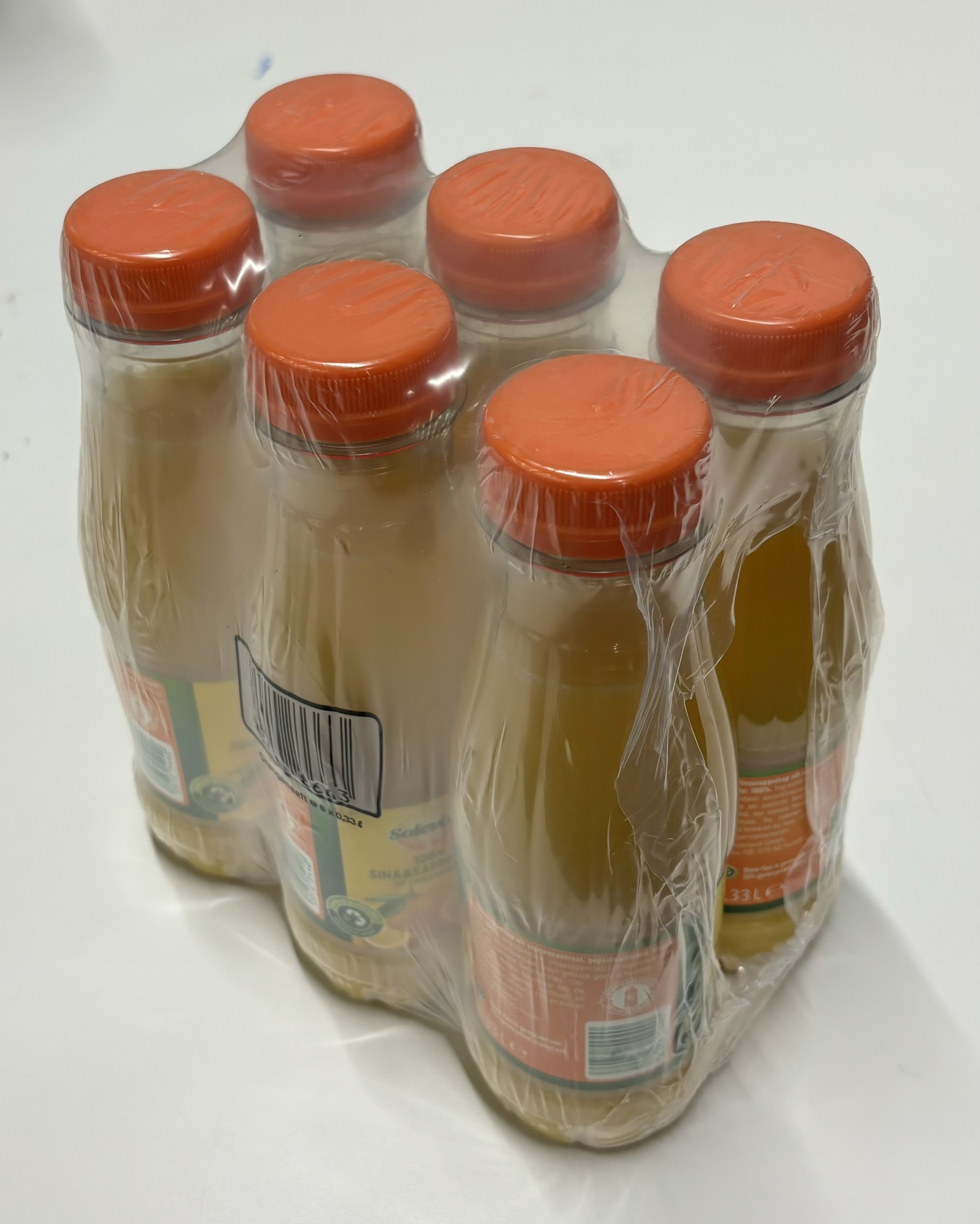}
			\caption{Juice bottles (2.12 kg)}
			\label{fig:snap_ref_post}
		\end{subfigure}
		\caption{Pictures of the different test objects used in the experimental validation.}
		\label{fig:objects}
	\end{figure*}	
    
    Out of the ten recorded references using the 1.25 kg rice box, three were selected that consistently produce successful experiments for all of the control approaches. On top of this, a single successful reference was recorded for three different grasping scenarios using 1) a single 0.64 kg parcel, 2) two parcels with a combined weight of 1.24 kg, and 3) a pack of plastic drink bottles weighing 2.12 kg. All of the test objects are depicted in Figure \ref{fig:objects}. 
    Experiments with each of the six references were then performed for five different initial object positions, with a y-displacement of -30 mm, -15 mm, 0 mm, 15 mm and 30 mm compared to the object position used during teleoperation. 
    Five experiments have been performed for the proposed control approach and the three baseline approaches for each object displacement value, resulting in a total of $6 \times 5 \times 5 \times 4 = 600$ additional experiments. 
    The main goal of these experiments is to evaluate the effectiveness of the approach in reducing the peaks and steps in the input forces.
    
    For the three references with the rice boxes, correct task execution is additionally evaluated by means of a motion capture system consisting of four Optitrack Prime x22 cameras, which records the position of the box during a grabbing cycle. Ideally, sustained contact between the box and the end effectors is established as soon as possible to reduce the risk of the box dropping, and to make sure the actual contact locations match the expected contact locations. The maximum lifting height of the box throughout a successful experiment indicates how quickly contact is established given the upward motion of the reference during impact. The higher the box reaches, the sooner contact is established. The average maximum lifting height is then determined for every combination of reference and y-displacement. For each experiment, the difference between the recorded maximum lifting height and the average for that reference and y-displacement is computed, and these differences are averaged for each of the four control approaches. 
    
    The results of these motion capture measurements are shown in the second column of Table \ref{tab:comparison}. From these results, it is clear that by far the lowest average maximum lifting height and hence the slowest contact establishment is recorded for the approach with no velocity feedback. This observation matches with the observations for a single experiment shown in Figure \ref{fig:results}, which shows a clear vibration in velocity when contact is established, and also matches with the large number of failed experiments depicted in the first column of Table \ref{tab:comparison}. The second lowest maximum height is recorded for the approach with no interim mode. This is likely caused by the premature transition to the post-impact mode, which results in a large velocity error that translates to a repulsive desired force away from the box, as shown in the third column of Figure \ref{fig:results}. 
    In the proposed approach, this repulsive drive is removed with the initial removal of velocity feedback while position feedback based on the ante-impact reference is retained. However, the gradual transition into the post-impact mode prevents the problems apparent from the approach with no velocity, which is reflected by the higher average maximum box height compared to both of these approaches. 
    The highest average box height is recorded for the approach with no RS, which is likely due to the nature of the uncertainty experiments, causing impacts to take place earlier than expected. Since the ante-impact velocity reference is followed until the nominal impact time, the velocity error initially pushes the end effectors further towards the box, causing faster contact establishment. However, this acceleration after the initial contact also causes a higher peak torque in the joints, increasing the amount of failure cases as previously mentioned and as seen in the first column of Table \ref{tab:comparison}.

\begin{figure}
		\centering
		\includegraphics[width=\linewidth]{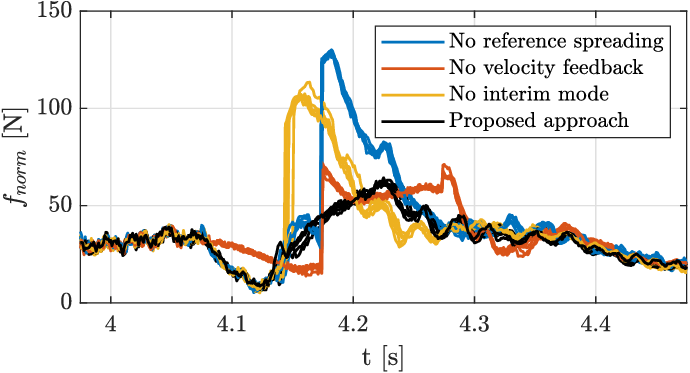}
		\caption{Desired force norm for the proposed control approach and baseline approaches, using a rice box with initial displacement of -30 mm.}
		\label{fig:Repeat_results_plot}
	\end{figure}

    To evaluate the success in reducing the peaks and steps in the input forces in each of the 600 experiments, we computed for every time step the norm of the desired force, i.e. 
    \begin{equation}
        f_\text{norm} = \left| \left[\vf_{1,x}^m, \vf_{1,y}^m, \vf_{1,z}^m, \vf_{2,x}^m, \vf_{2,y}^m, \vf_{2,z}^m \right]\right|
    \end{equation}
    with $m \in \{a,\text{int},p\}$ as the control mode at the corresponding timestep. 
    This desired force norm is plotted around the time of the impact event in Figure \ref{fig:Repeat_results_plot}. This plot depicts $f_\text{norm}$ for 20 experiments performed using the rice box for a single reference with an object displacement of -30 mm. First of all, it can be seen that the experiments are repeatable, as the desired force norms for the 5 experiments with the same control approach are highly similar over time. Second, in both the approach with no RS and the approach with no interim mode, a clear undesired jump in the desired force norm can be observed. For the approach with no RS, the largest jump occurs at the nominal impact time, as the post-impact mode is then activated regardless of the actual contact state. The largest jump for the approach with no interim mode occurs when the first impact is detected, since the post-impact mode is then activated before the impact event is completed. Meanwhile, no similar peaks or jumps of the desired force norm are observed for the proposed approach, again validating the approach. The severity of the force peak is, while still present due to the aforementioned discontinuity of the force feedforward signal, unsurprisingly also much lower for the approach with no velocity feedback. However, as shown from the results in Table \ref{tab:comparison}, this comes at the cost of control performance, while this is not the case for the proposed approach.

\begin{figure}		
		\centering
		\includegraphics[width=\linewidth]{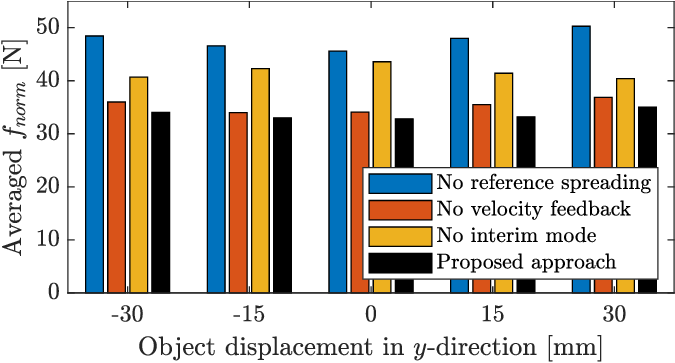}
		\caption{Average desired force norm around the nominal impact time for a total of 600 experiments with different objects and different initial displacements for each object, shown for the proposed control approach and the three baseline approaches.}
		\label{fig:Repeat_results}
    \end{figure}

    To further quantify the effect of reducing the input peaks, an average of $f_\text{norm}(t)$ is taken for each individual experiment over a 200 ms time frame centered around the nominal recorded impact time $T_r$. This $T_r$ can be recognized in Figure \ref{fig:Repeat_results_plot} as the time where the desired force norm in the approach with no reference spreading spikes up.    
    We then take another average of the obtained average force norms over the 30 experiments performed for each combination of displacement and control approach, obtaining a scalar value for each combination. 
    These averaged desired force norms are depicted in Figure \ref{fig:Repeat_results}. When looking at this diagram, it is apparent that the proposed control approach has the lowest desired force norm for each object displacement value, further highlighting the robustness of the proposed control approach in reducing input peaks and jumps under environmental uncertainty.
    
 
    One more thing that can be noticed is that even with 0 mm object displacement, the average desired force norms for the baseline approaches with no RS and no interim mode are high compared to the proposed approach, even though the impact between both end effectors and the object is happening simultaneously almost perfectly at the nominal impact time. The reason for this high force norm is that the compliance of the end effectors, the objects and the joint transmission causes the simultaneous impacts to occur over a finite time rather than at a single moment of time. This means that even in the absence of environmental uncertainty, still neither the ante-impact nor the post-impact reference can be reliably used for velocity feedback during this contact transition. Please note that using a fixed continuous reference would also not be a good solution, as the velocity during this transition can be hard to predict, while environmental uncertainty causes the time where the contact transition starts to be uncertain. 
    By removing velocity feedback at the beginning of the interim mode, the proposed approach is robust to this finite transition time, despite still having only discontinuous ante- and post-impact references at hand. When a near instantaneous impact event would be considered, it is expected the average desired force for the baseline approaches in the case of 0 mm displacement will shrink to the level of the proposed approach, while remaining larger than the proposed approach for increasing displacement values.

	
	\section{Conclusion}\label{sec:conclusion}
	
    This work has proposed and evaluated a reference spreading based QP robot control framework used for motions containing nominally simultaneous impacts in the context of dual-arm robotic grabbing. The aim of this framework is to accurately track a desired reference while avoiding impact-induced input peaks and steps. This is done by formulating overlapping ante- and post-impact references that are consistent with the impact dynamics, in this case generated through a novel teleoperation-based approach that uses impedance control with low gains. These references are used to generate a control action via three control modes: an ante-impact, interim and post-impact mode, each with a corresponding QP that uses high gains for accurate tracking. 
    
    Experimental validation for dual-arm grabbing of different objects shows the control framework is successful in reducing input peaks and steps compared to three baseline approaches while still successfully executing the task at hand. By artificially adding uncertainty in the object location, modifying the impact timing and thus causing a loss of simultaneity between the different impacts, we show the robustness of the proposed approach. 
    
    The key contributions of this particular work are the experimental evaluation of reference spreading on a complex dual arm robot setup, and a novel interim mode controller that avoids input peaks and jumps also during transitions between the interim- and post impact-phase without the need for a contact-completion detection mechanism that was advocated in previous reference spreading hybrid control literature. 
    
	A possible future extension is the formulation and inclusion of such a contact-completion detector into the control scheme, which could robustify the reference spreading framework to an even larger amount of uncertainty in the environment. Other future extensions are the inclusion of an impact-aware motion planner, removing the need for a user to demonstrate the desired motion using teleoperation, further investigation of impact detection in partially established contact conditions, and the formulation of a formal proof of stability for the proposed control approach.

	\addtolength{\textheight}{-0cm}  
	
	\section*{APPENDIX}

	\subsection{Derivation impedance control}\label{sec:impedance_derrivation}
	
	To show that forcing the error $\bm e_{i,\text{imp}}$ in \eqref{eq:e_imp_rec} to zero results in closed-loop behaviour that is identical to that of an impedance controller, we first rewrite the robot equations of motion in task space \cite{Khatib1987} by pre-multiplying \eqref{eq:eom} with $\bm J_i \bm M_i^{-1}$ to give
	\begin{equation}\label{eq:eom1}
	\bm J_i \ddot{\bm q}_i + \bm J_i\bm M_i^{-1}\bm h_i = \bm J_i\bm M_i^{-1} \bm \tau_i + \bm J_i\bm M_i^{-1} \bm J_{i}^T  \vf_i. 
	\end{equation}
	Adding and subtracting $\dot{\bm J}_i \dot{\bm q}_i$, we can rewrite \eqref{eq:eom1} as 
	\begin{equation}\label{eq:eom2}
	\bm J_i \ddot{\bm q}_i + \dot{\bm J}_i \dot{\bm q}_i - \dot{\bm J}_i \dot{\bm q}_i + \bm J_i\bm M_i^{-1}\bm h_i = \bm J_i\bm M_i^{-1} \bm \tau_i + \bm J_i\bm M_i^{-1} \bm J_{i}^T  \vf_i, 
	\end{equation}
	and eventually in the form 
	\begin{equation}\label{eq:eom3}
	\bm \Lambda_i \va_i + \bm \sigma_i = \vf_{i,\text{appl}} + \vf_i  
	\end{equation}
	with task-space inertia $\bm \Lambda_i := \left(\bm J_i\bm M_i^{-1} \bm J_{i}^T\right)^{-1}$, task space vector of Coriolis, centrifugal and gravitational terms $\bm \sigma_i := \bm \Lambda_i\left(\bm J_i\bm M_i^{-1}\bm h_i - \dot{\bm J}_i \dot{\bm q}_i\right)$, and $\bm \vf_{i,\text{appl}} := \bm \Lambda_i\bm J_i\bm M_i^{-1} \bm \tau_i$. 
	
	Following the same procedure outlined to obtain \eqref{eq:eom3} we can also convert \eqref{eq:EOM_free}, which is used to transform the desired joint acceleration into a desired input torque, to task-space, which results in 
	\begin{equation}\label{eq:appl_wrench1}
	\vf^*_{i,\text{appl}} = \bm \Lambda_i \va_i^* + \bm \sigma_i,
	\end{equation}
	where $\vf^*_{i,\text{appl}}$ denotes the input wrench, and $\va_i^*$ denotes the desired end effector acceleration. Convergence of $\bm e_{i,\text{imp}}$ from \eqref{eq:e_imp_rec} to zero implies that 
    \begin{equation}\label{eq:va_i_app}
        \va^*_i = \bm \Lambda_i^{-1} \vf_{i,r}.
    \end{equation}
    Combining \eqref{eq:imp_des}, \eqref{eq:appl_wrench1} and \eqref{eq:va_i_app}, the applied wrench $\vf^*_{i,\text{appl}}$ from \eqref{eq:appl_wrench1} is given by
	\begin{equation}\label{eq:appl_wrench2}
	\vf^*_{i,\text{appl}} = \bm D_{r} \left( \vv_{i,d} - \vv_i \right) + \bm K_{r} \begin{bmatrix}\bm p_{i,d} - {\bm p_{i}} \\ \bm R_i(\log(\bm R_i^T{\bm R}_{i,d}(t)))^{\vee } \end{bmatrix} + \bm \sigma_i,
	\end{equation}
	which is indeed corresponding to traditional impedance control with compensation of the gravity, Coriolis and gravitational terms. 
	Substituting \eqref{eq:appl_wrench2} into \eqref{eq:eom3} results in the desired task-space closed-loop behaviour
	\begin{equation}\label{eq:eom_closed}
	\bm \Lambda_i \va_i - \bm D_{r} \left( \vv_{i,d} - \vv_i \right) - \bm K_{r} \begin{bmatrix}\bm p_{i,d} - {\bm p_{i}} \\ \bm R_i(\log(\bm R_i^T{\bm R}_{i,d}(t)))^{\vee } \end{bmatrix} = \vf_i, 
	\end{equation}
	which does resemble a mass-spring-damper system where the end effector is attached to the target reference, indeed corresponding to traditional impedance control.

    \subsection{Impact detection}\label{sec:detection}
	
	Since both the reference extension procedure in Section \ref{sec:reference_extension} as well as the control approach in Section \ref{sec:control_approach} rely on detection of the first moment of impact, an impact detection approach has been formulated that is used in both these instances. Please note that this procedure can be replaced by any other impact detection approach with equal performance without requiring any modification to the approaches presented in Section \ref{sec:reference_generation} and \ref{sec:control_approach}. 
	
	Since we do not use a force sensor, the signals at hand are the end effector velocity $\bm v_i(t)$ and an estimation of the externally applied force $\bm f_{i,\text{est}}(t)$ using a momentum observer \cite{DeLuca2005}, in our use case provided by the Franka Control Interface. Upon an impact of either robot with robot index $i \in \{1,2\}$, a sudden change should be apparent in both signals. However, solely considering sudden changes in $\bm v_i(t)$, similar to approaches like \cite{Rijnen2018a}, could lead to false positives in impact detection with a flexible joint robot when an aggressive control strategy is used while the robot is in free motion. Meanwhile, only considering rapid changes in $\bm f_{i,\text{est}}(t)$ could lead to false positives when the robot is already in contact, and an aggressive control strategy causes a sudden increase in the contact force. Hence, our proposed impact detection approach considers both signals. We identify an impact has happened when the following three conditions hold for a given time $t$:
	\begin{enumerate}
		\item $\left \| \bm f_{i,\text{est}}(t-\Delta t_\text{det}) \right \|< b_{f,\text{low}}$,
		\item $\left \| \bm f_{i,\text{est}}(t) \right \|> b_{f,\text{high}}$,
		\item $\bm v_i(t-\Delta t_\text{det}) \cdot \bm f_{i,\text{est}}(t) < -b_v\left \| \bm f_{i,\text{est}}(t) \right \|$,
	\end{enumerate}
	for a pre-defined time interval $\Delta t_{\text{det}}$ and pre-defined force and velocity bounds $b_{f,\text{low}}, b_{f,\text{high}}, b_v \in \mathbb{R}^+$, with $b_{f,\text{high}} > b_{f,\text{low}}$. The parameter values used for the experiments in this work are $b_{f,\text{low}} = 4$N, $b_{f,\text{high}} = 8$N, $b_v = 0.025$m/s and $\Delta t_{\text{det}} = 0.2$s.
 
    The interpretation of conditions 1 and 2 is that a sudden increase in the estimated force must be observed in a short time frame $\Delta t_\text{det}$. 
	Condition 3 implies that before the suspected impact, a significantly large velocity in opposite direction of the estimated contact force is measured, removing the false positive that appears when a sudden increase of the contact force is measured while contact is already established. The first time $t$ for which each condition holds for any of the two robots is marked as the initial impact time, called $T_r$ in the recording procedure in Section \ref{sec:reference_generation}, and $T_\text{imp}$ in the autonomous control approach in Section \ref{sec:control_approach}.
 

	\bibliography{References/library}{}
	\bibliographystyle{ieeetr}

\begin{IEEEbiography}[{\includegraphics[width=1in,height=1.25in,clip,keepaspectratio]{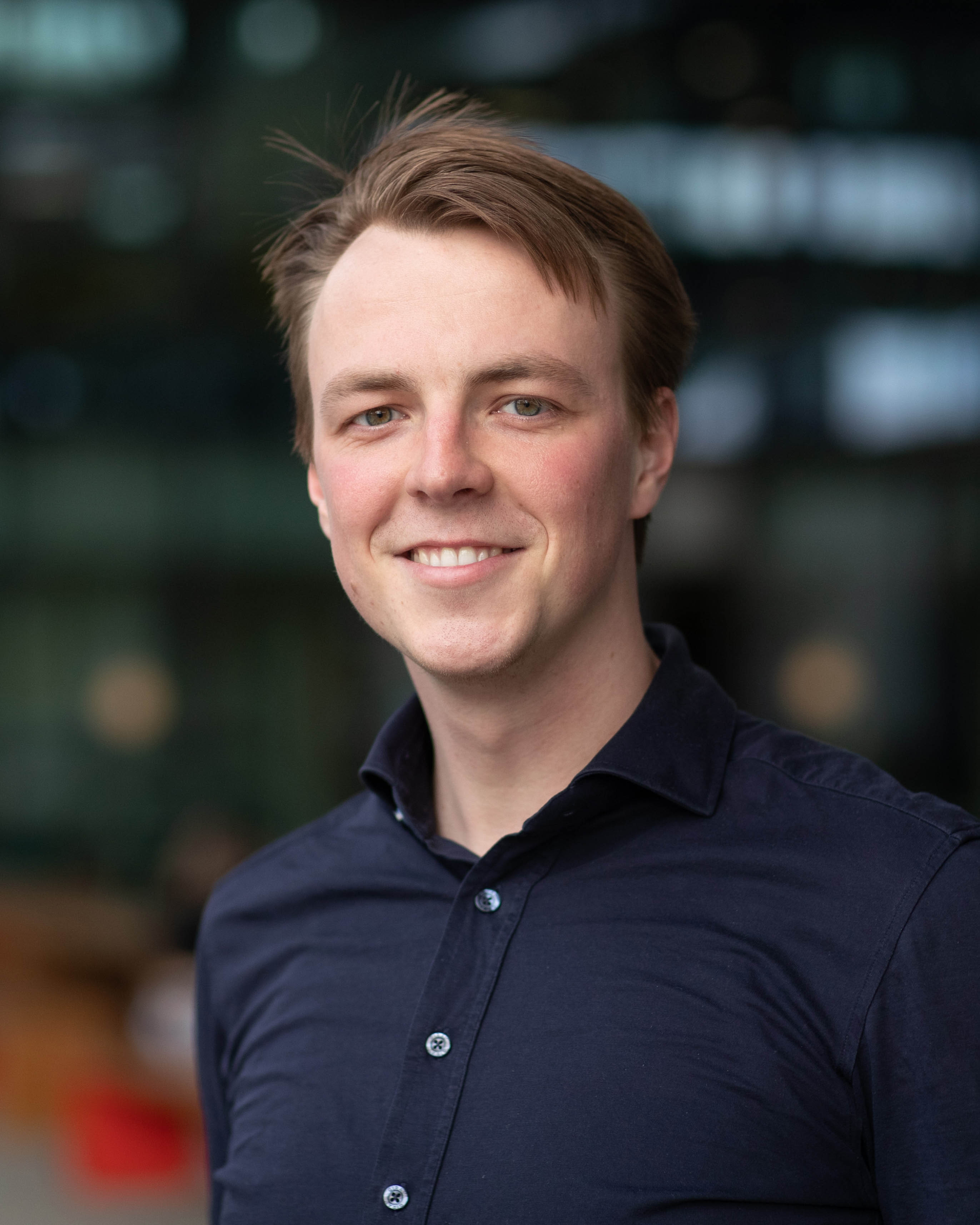}}]{Jari van Steen} obtained his master’s degree in Mechanical Engineering from the Eindhoven University of Technology, the Netherlands, in 2020. Later that year, he started as a PhD student at Eindhoven University of Technology within the European Horizon 2020 project I.AM. This project aims to exploit impacts in robotics with the aim of improving applicability of robots in logistic settings, such as automated depalletization. His main research focus is on impact-aware control in robotic manipulation. 
\end{IEEEbiography}

\vspace{-24pt}

\begin{IEEEbiography}[{\includegraphics[width=1in,height=1.25in,clip,keepaspectratio]{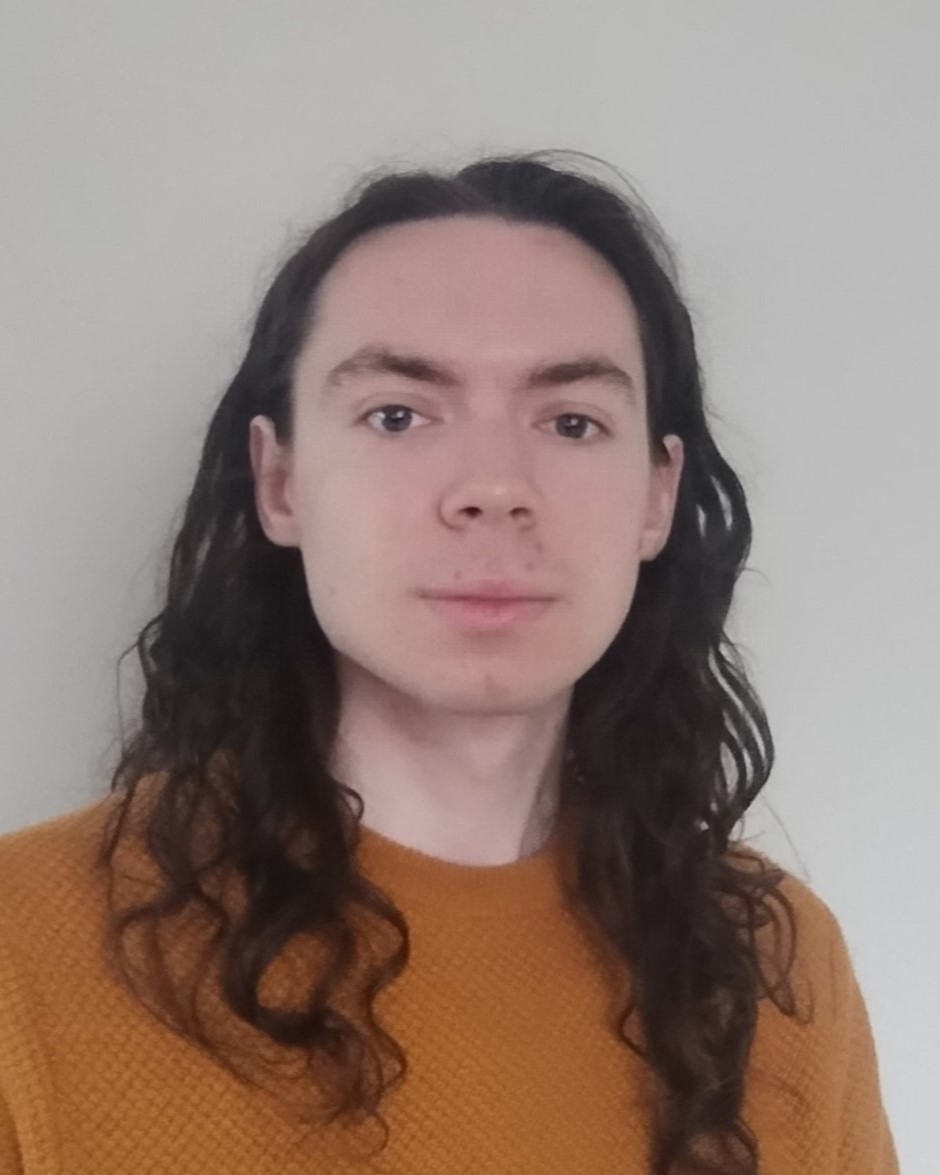}}]{Gijs van den Brandt} is a PhD candidate in the Robotics Group at Eindhoven University of Technology's Mechanical Engineering Department. His research focuses on world modeling for robotic manipulation, aiming to improve robots' adaptability in volatile manufacturing environments. Gijs earned his bachelor's degree in 2020 and his master's degree in 2023, both in mechanical engineering from Eindhoven University of Technology. His graduation thesis involved the experimental validation of an impact-exploiting control method for robotic manipulation in logistics. 
\end{IEEEbiography}

\vspace{-24pt}

\begin{IEEEbiography}[{\includegraphics[width=1in,height=1.25in,clip,keepaspectratio]{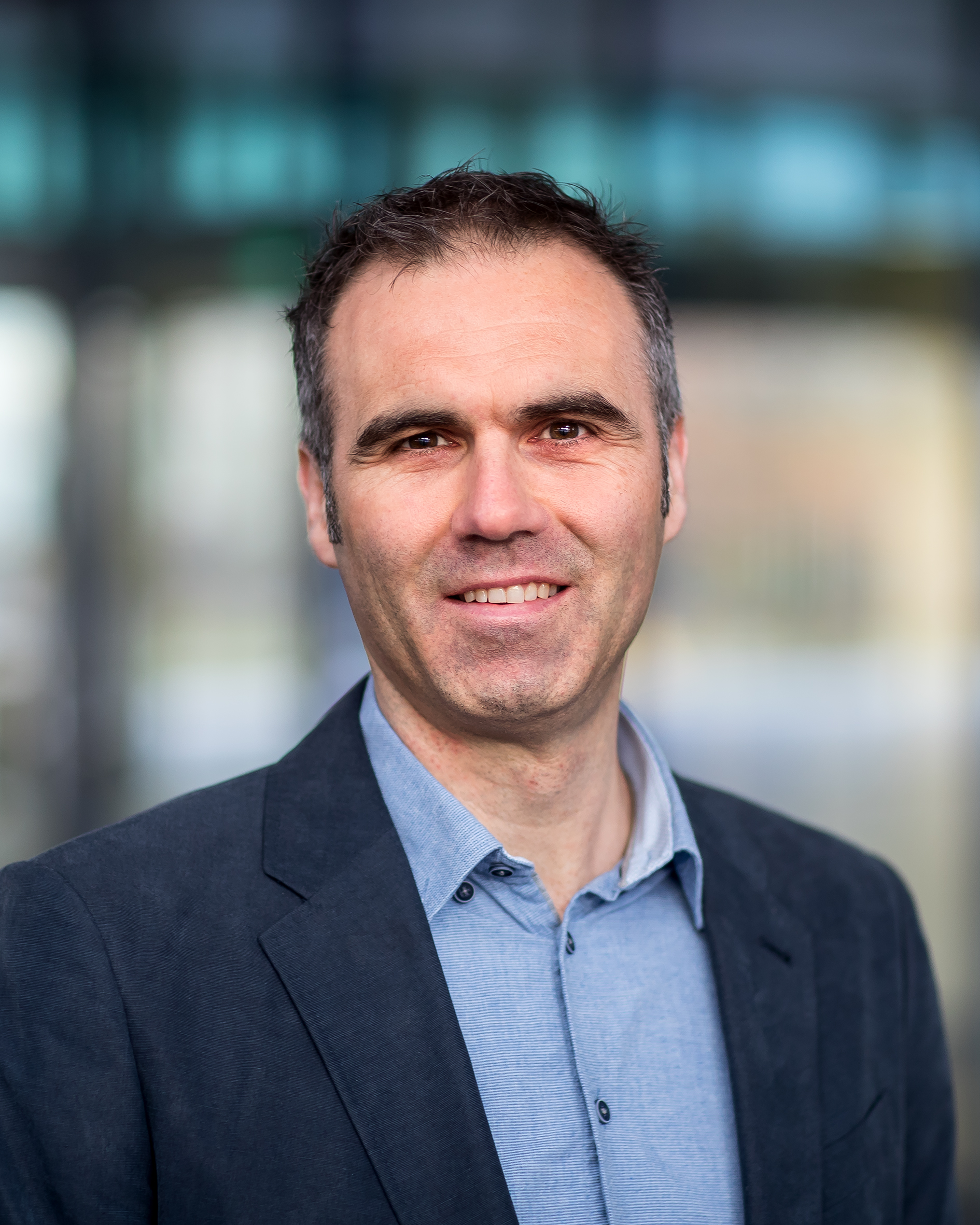}}]{Nathan van de Wouw} obtained his Ph.D.-degree in Mechanical Engineering from the Eindhoven University of Technology (TU/e), the Netherlands, in 1999. He holds a full professor position at the Mechanical Engineering Department of TU/e. 
He has held a (part-time) full professor position the Delft University of Technology, the Netherlands (2015-2019), and an adjunct full professor position at the University of Minnesota, U.S.A. (2014-2021). In 2015, he received the IEEE Control Systems Technology Award ``For the development and application of variable-gain control techniques for high-performance motion systems''. He is an IEEE Fellow for his contributions to hybrid, data-based and networked control.
\end{IEEEbiography}

\vspace{-24pt}

\begin{IEEEbiography}[{\includegraphics[width=1in,height=1.25in,clip,keepaspectratio]{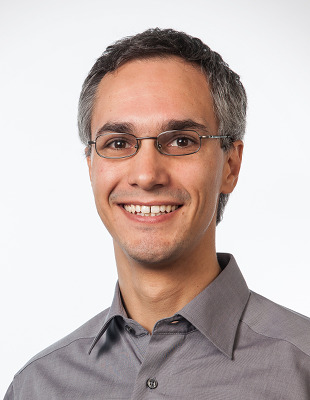}}]{Jens Kober} (Senior Member, IEEE)  received the Ph.D. degree in engineering from TU Darmstadt, Darmstadt, Germany, in 2012.  
He is currently an Associate Professor with TU Delft, Delft, The Netherlands. He was a Postdoctoral Scholar jointly with CoR-Lab, Bielefeld University, Bielefeld, Germany, and with Honda Research Institute Europe, Offenbach, Germany.     
Dr. Kober was a recipient of the annually awarded Georges Giralt PhD Award for the best PhD thesis in robotics in Europe, the 2018 IEEE RAS Early Academic Career Award, the 2022 RSS Early Career Award, and was a recipient of an ERC Starting grant.
\end{IEEEbiography}

\vspace{-24pt}

\begin{IEEEbiography}
[{\includegraphics[width=1in,height=1.25in,clip,keepaspectratio]{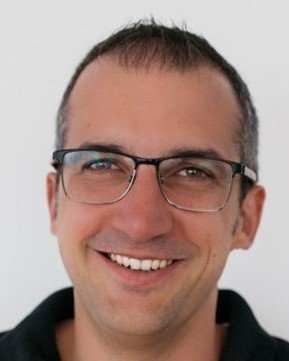}}]{Alessandro Saccon} is Associate Professor at the Mechanical Engineering Department of the Eindhoven University of Technology (TU/e). His areas of expertise include robotics, nonlinear control, numerical optimal control, multi-body dynamics, geometric mechanics, and computer vision. His current research efforts are directed  toward the development and validation of innovative control strategies for robotic systems  with multiple intermittent dynamic contacts, with application in the field of contact-rich and impact-aware robot manipulation (coordinator of the H2020 EU project I.AM.). He received a laurea degree cum laude in computer engineering and a Ph.D. in control system theory from the University of Padua, Italy.
\end{IEEEbiography}

\end{document}